\documentclass[11pt]{article}
\usepackage[T1]{fontenc}
\usepackage[utf8]{inputenc}
\usepackage{amsmath,amsfonts,amsthm,amssymb}
\usepackage{setspace}
\usepackage{fullpage}
\usepackage{fancyhdr}
\usepackage{xspace, xcolor}

\usepackage{verbatim}
\usepackage{lastpage}
\usepackage{natbib}

\usepackage{titletoc}

\usepackage{times}
\usepackage{setspace}
\usepackage{graphicx,float,wrapfig}
\usepackage{algorithm}
\usepackage{enumitem}
\usepackage[noend]{algpseudocode}
\usepackage{subcaption}
\allowdisplaybreaks

\newtheorem{theorem}{Theorem}[section]
\newtheorem{lemma}[theorem]{Lemma}

\newtheorem{proposition}[theorem]{Proposition}

\newtheorem{cond}{Condition}
\theoremstyle{definition}
\newtheorem{fact}[theorem]{Fact}

\def\shownotes{0}  %
\ifnum\shownotes=1
\newcommand{\authnote}[2]{[#1: #2]}
\else
\newcommand{\authnote}[2]{}
\fi

\usepackage{hyperref}
\hypersetup{
	colorlinks   = true, %
	urlcolor     = red, %
	linkcolor    = blue, %
	citecolor   = blue %
}
\usepackage{cleveref}

\usepackage{amsmath,amsfonts,bm}

\newcommand*{\defeq}{\triangleq}

\def\1{\bm{1}}

\makeatletter
\newcommand{\ve}{\@ifnextchar\bgroup{\velong}{{\bm{e}}}}
\newcommand{\velong}[1]{{\bm{#1}}}
\makeatother

\DeclareMathAlphabet{\mathsfit}{\encodingdefault}{\sfdefault}{m}{sl}
\SetMathAlphabet{\mathsfit}{bold}{\encodingdefault}{\sfdefault}{bx}{n}

\def\calC{{\mathcal{C}}}
\def\calD{{\mathcal{D}}}

\def\calF{{\mathcal{F}}}

\def\calH{{\mathcal{H}}}

\def\calL{{\mathcal{L}}}

\def\calN{{\mathcal{N}}}

\newcommand{\E}{\mathbb{E}}
\newcommand{\R}{\mathbb{R}}

\newcommand{\KL}{D_{\mathrm{KL}}}

\DeclareMathOperator*{\argmax}{argmax}
\DeclareMathOperator*{\argmin}{argmin}

\DeclareMathOperator{\sign}{sign}

\def\({\left(}
\def\){\right)}
\def\[{\left[}
\def\]{\right]}
\newcommand{\poly}{\mathrm{poly}}
\newcommand{\polylog}{\mathrm{polylog}}

\newcommand{\dotp}[2]{\left<#1, #2\right>}

\newcommand{\norm}[1]{\left\| #1 \right\|}

\newcommand{\relu}{\mathrm{ReLU}}

\newcommand{\abs}[1]{{\left| {#1} \right|}}

\newcommand{\ind}[1]{\mathbb{I}\left[ #1 \right]}

\newcommand{\dd}{\mathrm{d}}
\newcommand{\TV}{D_{\rm TV}}

\newcommand{\Pkdb}{\bar{P}_{k,d}}
\newcommand{\Pkd}{P_{k,d}}
\newcommand{\Pldb}{\bar{P}_{l,d}}

\newcommand{\Nkd}{N_{k,d}}
\newcommand{\Nld}{N_{l,d}}
\newcommand{\Ykd}{\mathbb{Y}_{k,d}}

\newcommand{\Sp}{\mathbb{S}^{d-1}}

\newcommand{\nnrelu}{\mathrm{NN}\text{-}\mathrm{ReLU}}
\newcommand{\nnexp}{\mathrm{NN}\text{-}\mathrm{Exp}}
\newcommand{\mud}{{\mu_d}}
\newcommand{\Proj}{\Pi}

\renewcommand{\Pr}{\mathop{\rm Pr}\nolimits}
\renewcommand{\tilde}{\widetilde}

\author{Kefan Dong \\ 
	Stanford University \\
	\texttt{kefandong@stanford.edu}
	\and
	Tengyu Ma \\
	Stanford University \\
	\texttt{tengyuma@stanford.edu}
}

\title{Toward $L_\infty$-recovery of Nonlinear Functions: A Polynomial Sample Complexity Bound for Gaussian Random Fields}

\begin{document}
	\maketitle
	
	\begin{abstract}
		Many machine learning applications require learning a function with a small worst-case error over the entire input domain, that is, the $L_\infty$-error, whereas most existing theoretical works only guarantee recovery in average errors such as the $L_2$-error.  
		$L_\infty$-recovery from polynomial samples is even impossible for seemingly simple function classes such as constant-norm infinite-width two-layer neural nets. This paper makes some initial steps beyond the impossibility results by leveraging the randomness in the ground-truth functions. We prove a polynomial sample complexity bound for random ground-truth functions drawn from Gaussian random fields. Our key technical novelty is to prove that the degree-$k$ spherical harmonics components of a function from Gaussian random field cannot be spiky in that their $L_\infty$/$L_2$ ratios are upperbounded by $O(d \sqrt{\ln k})$ with high probability. In contrast, the worst-case $L_\infty$/$L_2$ ratio for degree-$k$ spherical harmonics is on the order of $\Omega(\min\{d^{k/2},k^{d/2}\})$. 

	\end{abstract}
	
	\section{Introduction}\label{sec:intro}

Classical statistical learning theory primarily concerns with recovering functions from examples  with small errors \textit{averaged} over a distribution of inputs, e.g., the mean-squared error (that is, the $L_2$-error with respect to the test distribution).
However, the worst-case error over the entire input domain, that is, the $L_\infty$-error, is crucial for many applications, and also challenging to achieve. 
For example, an $L_\infty$-error recovery guarantee will make the learned function more robust to adversarial examples, while standard training is vulnerable~\citep{goodfellow2015explaining,madry2017towards}. 
The $L_\infty$-recovery is also necessary for many applications where the recovered models will be further used in a downstream decision making process, such as model-based bandits~\citep{huang2021optimal}, reinforcement learning~\citep{huang2021going,sutton2018reinforcement}, and physics informed neural networks \citep{raissi2019physics,wang20222}. In particular, recent theoretical works on deep reinforcement learning heavily rely on the $L_\infty$-recovery of the $Q$-function to prevent the actions from misusing a small worst-case region of inputs where the error is much larger than the average error~\citep{huang2021going}.  (See Section~\ref{sec:relatedwork} for more discussions on the applications.)

This paper focuses on $L_\infty$-error recovery of nonlinear functions from polynomial samples. For a compact domain $D$, we aim to learn a function that is \emph{pointwise} close to the ground-truth over the entire domain $D$. Formally, given polynomial random input-output pairs from a ground-truth function $f$, our goal is to learn a function $g$ with small $L_\infty$-distance/error to $f$, defined by $$\|f-g\|_\infty\defeq\sup_{x\in D}|f(x)-g(x)|.$$

For linear function class, we can straightforwardly $L_\infty$-recovery guarantees by relating the $L_\infty$-error to the $L_2$-error on the test distribution, which in turn can be bounded standard tools such as uniform convergence~(e.g., \citet{bartlett2002rademacher,koltchinskii2002empirical,wei2019improved}). Concretely, suppose $P$ is the training/test distribution on domain $D$ and the covariance matrix $\Sigma =\E_{x\sim P}[xx^\top]$ is full-rank,  we have for any linear functions $f$ and $g$, 
\begin{align}
\textstyle{\|f-g\|_\infty\le \left(\sup_{x\in D}\|x\|_2 \right) \cdot \lambda_{\min}(\Sigma)^{-1/2} \cdot \|f-g\|_{L_2(P)} \,.} \label{eqn:2}
\end{align}
where $\lambda_{\min}(\Sigma)$ is the minimum eigenvalue of $\Sigma$ and $\|f-g\|_{L_2(P)} = \left(\E_{x\sim P} (f(x)-g(x))^2\right)^{1/2}$ is the squared error on the distribution $P$. Therefore, we can reduce the $L_\infty$-recovery to $L_2$-recovery, and the inequality above is  tight for most scenarios. 

In contrast, $L_\infty$-recovery of nonlinear functions is much more challenging. 
When the model's parameters are identifiable and can be recovered, e.g., for finite-width two-layer neural nets (without biases)~\citep{zhong2017recovery,zhou2021local} or low-degree polynomials \citep{huang2021going}, $L_\infty$-recovery of the functions follows straightforwardly from parameter recovery. 
Parameter recovery fundamentally requires the sample size to be larger than parameter dimension, and therefore does not apply to the over-parameterized settings that are ubiquitous in modern machine learning (\citet{zagoruyko2016wide,du2018gradient,allen2019learning,zhang2021understanding} and references therein) or infinite dimensional features in the kernel method settings.
Existing $L_\infty$-recovery algorithms require quasi-polynomial or exponential in dimension samples for two-layer neural networks~\citep{mhaskar2006weighted,mhaskar2019function} or general very smooth functions (with decaying higher-order derivatives)~\citep{vybiral2014weak,krieg2019uniform} .

In fact, $L_\infty$-recovery from polynomial samples is impossible for even seemingly simple function classes, such as two-layer single-neuron neural nets with bias~\citep{dong2021provable,li2021eluder} or constant-norm infinite-width two-layer neural nets without bias (Theorem~\ref{thm:relu-lb} of this paper).
The fundamental challenge is that these function classes contain many \emph{spiky} functions $f$ such that $\|f\|_2 \ll \|f\|_\infty$, which means that inequalities analogous to Eq.~\eqref{eqn:2} cannot hold. Moreover, these functions may mostly have tiny values except a spike on an exponentially-small region. Likely, none of the polynomial number of examples falls into the spiky region. As a result, the spike cannot be identified, and $L_\infty$-recovery cannot be achieved.

Interestingly, $L_\infty$-recovery of functions in reproducing kernel Hilbert space (RKHS) with polynomial samples is still a challenging open question, even though $L_2$-recovery with polynomial samples and time has been well established \citep{bartlett2002rademacher,hofmann2008kernel}.
Even though they are essentially linear functions with an infinite dimensional features, analysis analogous to linear models (e.g., Eq~\eqref{eqn:2}) is vacuous because the covariance of the features, that is the kernel function, typically has a sequence of eigenvalues that decays to zero.
In fact, $L_\infty$-recovery of functions with constant RKHS norm for various kernels (e.g., the radial basis function (RBF) kernels kernel) requires exponential number of samples~\citep{scarlett2017lower,kuo2008multivariate}. Intuitively, this is because RKHS still contains spiky functions, e.g., 
the $k$-th eigenfunctions of the RBF kernel (or any inner product kernel) on the unit sphere can be spiky for relatively large $k$.

Towards going beyond these intractability results and achieve polynomial sample complexity bounds,  we make additional randomized and smooth assumptions on the ground-truth functions that we aim to recover.  We essentially assume that the ground-truth function has decaying and random high-frequency components.

Concretely, we work with random ground-truth functions $f$ drawn from a Gaussian random field (GRF, also known as Gaussian process) on the unit sphere \citep{seeger2004gaussian,lang2015isotropic}. We assume that the covariance (or kernel) function, denoted by $K:\Sp\times\Sp\to\R$, is an inner product function given by $K(x,x')=\kappa(x^\top x')$ for some function $\kappa:[-1,1]\to\R$, which means that the GRF is isotropic.

All the inner product kernels $\kappa(x^\top x')$ on the unit sphere share the same eigenfunctions called spherical harmonics \citep{atkinson2012spherical}. 
This brings opportunities for us to use the spherical harmonics tools to analyze the problem. 
Spherical harmonics form a complete set of basis for the square integrable functions over the sphere (in analog to the Fourier basis in $\R^d$). 
Intuitively, fast decay of the spherical harmonics components of the function implies the function is smoother. Moreover, some higher-degree spherical harmonics can be more spiky and challenging to recover in $L_\infty$-error.

Our main result is an $L_\infty$-recovery algorithm (Alg.~\ref{alg:main}) with polynomial sample complexity for a random ground-truth $f$ drawn from Gaussian random fields, given that the $k$-th eigenvalues of the covariance function $K$ decays at a rate $O(k^{-(1+\alpha) d})$ for any universal constant $\alpha>0$ (Theorem~\ref{thm:mainsc}). This decay rate is equivalent to that the degree-$k$ spherical harmonics component of $f$ is on the order $O(k^{-\alpha d/2})$. We note that the randomness from the Gaussian random field is the key for us to work with this decay (that is, $\alpha > 0$), because a worst-case function with $O(k^{-d/2})$ decay in the spherical harmonics components is impossible to recover with polynomial samples (Lemma~\ref{lem:relu-lb}). 
This suggest that the randomness property in the ground-truth function makes the $L_\infty$-recovery much easier. 
Moreover, for comparison, $L_2$-recovery with polynomial sample is only possible when $\alpha \ge 0$.\footnote{The $\alpha = 0$ case is subtle for both $L_2$- and $L_\infty$-recovery and we leave it as an open question for future work. } 
In other words, the randomness assumption qualitatively makes $L_\infty$-recovery as easy as $ L_2$-recovery.

Our main technique is to prove a much tighter upper bound for the $L_\infty$-norm of the high-degree components of the ground-truth $f$ using the randomness. Lemma~\ref{lem:GRF} shows that the high-degree components of $f$ drawn from Gaussian random processes are \emph{not} spiky: their $L_\infty$/$L_2$ ratios are upper bounded by $O(d\sqrt{\ln k})$ with high probability, whereas the worst-case ratio is $\Omega(\min\{d^{k/2},k^{d/2}\})$. 
This lemma might be of independent interest as it extends \citet[Theorem 1]{burq2014probabilistic} to the high-dimensional case with a precise bound on the dependency on $d$. It was not known that the dependency on $d$ is polynomial. 
Our proof is also surprisingly much simpler than that in~\citet{burq2014probabilistic}. 
Hence, when the eigenvalues of kernel $K$ decays, we can truncate $f$ at degree $\tilde{O}(1)$ and get a low-degree polynomial approximation with small $L_\infty$-error (Lemma~\ref{lem:main}).

The rest of this paper is organized as follows. Section~\ref{sec:relatedwork} discusses additional related works and the applications of $L_\infty$-recovery to bandits and reinforcement learning problems. In Section~\ref{sec:preliminary} we give a concise overview of the spherical harmonics, the important tools in this paper. Section~\ref{sec:main-results} states our algorithm and proves a polynomial sample complexity bound for recovering random functions from a Gaussian random field. \Cref{sec:lowerbounds} proves that $L_\infty$-recovery is impossible for two-layer neural nets without bias (Theorem~\ref{thm:relu-lb}), which may be of independent interest.

\paragraph{Additional notations.} Let $\Sp$ be the $(d-1)$-dimensional unit sphere.
We assume that the training distribution is uniform over the $\Sp$. 
That is, the data $x_i$ is sampled independently and uniformly from the sphere, and $y_i=f(x_i)+\calN(0,1)$ where $y$ is the ground-truth.
With slight abuse of notations, we also use $\Sp$ to denote the uniform distribution over the unit sphere. For a function $g:\Sp\to\R$, let $\|g\|_p\defeq \E_{x\sim \Sp}[g(x)^p]^{1/p}$ be its $L_p$-norm with respect to the uniform distribution. For two functions $g,h:\Sp\to\R$, $\dotp{g}{h}\defeq \E_{x\sim \Sp}[h(x)g(x)]$ denotes their inner product. For a function $h:\R\to\R$, we use $h^{(k)}$ to denote its $k$-th derivative.

In the following, for two non-negative sequences $a_{k},b_{k}$, we write $a_k=O(b_k)$ or $a_k\lesssim b_k$ if there exists an \emph{absolute} constant $c$ such that $a_k\le c b_k$ for every $k\ge 0$. We write $a_k=\tilde{O}(1)$ if $a_k=O(\polylog(k))$, and $a_k=\Theta(b_k)$ if $a_k\lesssim b_k$ and $b_k\lesssim a_k.$

	\section{Related Works}\label{sec:relatedwork}
\sloppy The classical uniform convergence framework (e.g., \citet{bartlett2002rademacher,koltchinskii2002empirical,kakade2009complexity,bartlett2017spectrally,wei2019improved}) does not directly solve the $L_\infty$-recovery problem. This is because for any $p>0$, approximating the $L_p$-error (defined by $\|f-g\|_p\defeq\E_{x\sim D}[|f(x)-g(x)|^p]^{1/p}$) with $\epsilon$ precision requires $\poly(\epsilon^{-p})$ samples. Hence, as $p\to\infty$, we cannot avhieve uniform convergence using polynomial samples, meaning that bounding the $L_\infty$-error of the learned function requires novel analysis.

\paragraph{Gaussian process bandits and kernelized bandits.} A closely related line of research is the Gaussian process bandits. Instead of learning a function with small $L_\infty$-error, Gaussian process bandit algorithms aim to find a $x$ that maximizes the function $f(x)$ when $f$ is drawn from a Gaussian process \citep{grunewalder2010regret}.
Most of the existing results focuses on radial basis function kernel and Mat\'ern kernels, and the regret is exponential in the ambient dimension $d$ \citep{srinivas2009gaussian,krause2011contextual,shekhar2018gaussian,vakili2021scalable}. 
For Gaussian processes with non-isotropic kernels, \citet{grunewalder2010regret} prove exponential regret upper and lower bounds.
For general kernels, \citet{lederer2019uniform,lederer2021uniform} proves a $L_\infty$-error bounds for Gaussian process regression with no assumption on the spectrum of the covariance, and the sample complexity is also exponential.

Another line of research focuses on kernelized bandits and assumes that ground-truth $f$ has a small RKHS norm (see \citet{valko2013finite,wang2014theoretical,chowdhury2017kernelized,vakili2021information,zhang2021neural} and references therein). For the RBF and Mat\'ern kernels, their regret bounds are exponential in $d$ and \citet{scarlett2017lower} prove that no algorithm can achieve polynomial sample complexities. Since a small RKHS norm does not exclude spiky functions in general, the results in this setting requires a stronger assumption on ground-truth $f$. In fact, a function drawn from a Gaussian process has a infinite RKHS norm (defined by the same kernel) almost surely \citep{wahba1990spline}. 

\paragraph{Neural nets recovery.}
The parameters of finite-width two-layer neural networks can be recoverd with additional assumptions on the correlation between neurons \citep{zhong2017recovery,fu2020guaranteed,zhou2021local}, or the condition number of the first-layer weights \citep{zhang2019learning}. For two-layer neural networks with unbiased ReLU activation, \citet{bakshi2019learning} design algorithms whose sample complexity scales exponentially in the number of hidden neurons.
In addition, \citet{milli2019model} proves a recovery guarantee of two-layer ReLU neural networks when the algorithm can query the gradient of the ground-truth. 
These methods cannot be applied to infinite-width neural networks, and $L_\infty$-recovery of a infinite-width neural network requires exponential samples (Theorem~\ref{thm:relu-lb}).

\paragraph{Applications to bandits, reinforcement learning, and PINN.} The best-arm identification problem in nonlinear bandits can be reduced to an $L_\infty$-recovery problem. If we can learn a function $g$ that approximates the true reward $f$ with $\|f-g\|_\infty\le \epsilon/2$, the action $\hat{x}\defeq \argmax_{x\in D}g(x)$ is $\epsilon$-optimal.
Similarly, if the $Q$-function can be learned with a small $L_\infty$-error for finite horizon reinforcement learning, we can guarantee the optimality of the learned policy \citep{huang2021going}.

For physics informed neural networks \citep{raissi2019physics}, minimizing the $L_2$ loss may not be satisfactory \citep{wang20222,krishnapriyan2021characterizing,wang2021understanding}. When learning the Hamilton-Jacobi-Bellman equations (an analog of Bellman equations for continuous time), \citet{wang20222} proves that a small $L_\infty$-error can guarantee a good final performance, while a small $L_2$-error cannot.

\paragraph{$L_\infty$-recovery for other nonlinear functions.} Several other related works study the $L_\infty$-recovery with different assumptions on the ground-truth function. \citet{bertin2004minimax,korostelev1994asymptotically,tsybakov1998pointwise,golubev2000adaptive} study the minimax rate for $L_\infty$-recovering for one-dimensional smooth functions (e.g., functions in H\"older, Sobolev, or Besov classes).
For general functions with bounded or decaying high-order derivatives, \citet{vybiral2014weak,krieg2019uniform} design estimators with quasi-polynomial or exponential in dimension samples. \citet{ibragimov1984asymptotic,stone1982optimal,nyssbaum1987nonparametric,bertin2004asymptotically} determine the asymptotically optimal rate for $L_p$-recovery of H\"older smooth functions for general $p\in [1,\infty]$ in high dimensions.

When the ground-truth function lies in the reproducing kernel Hilbert space, \citet{kuo2009power} prove some sufficient conditions for $L_\infty$-recovery with polynomial sample complexity. In general, $L_\infty$-recovery with polynomial samples is impossible unless the eigenvalues of the kernel decay very fast \citep{long2023reinforcement, kuo2008multivariate}.
We refer the readers to \citet{ebert2021tractability} for a comprehensive survey in this direction.

Another line of research focuses on learning a nonlinear function with respect to the Sobolev norm \citep{fischer2020sobolev,steinwart2009optimal}. While their analysis can lead to $L_\infty$-recovery bounds, they require stronger smoothness assumptions to exclude the worst-case hard instances shown in  Lemma~\ref{lem:relu-lb}. In contrast, our algorithms achieve $L_\infty$-recovery in the average case using much weaker smoothness assumptions.

\section{Preliminaries on Spherical Harmonics}\label{sec:preliminary}
Now we give a brief overview of spherical harmonics, the essential tools in this paper, based on \citet[Section 2]{atkinson2012spherical}. Spherical harmonics are the eigenfunctions of the Laplacian operator on the sphere. The eigenfunctions corresponding to the $k$-th eigenvalue are degree-$k$ polynomials, and form a Hilbert space denoted by $\Ykd.$ The dimension of $\Ykd$ is $\Nkd\defeq \binom{d+k-1}{d-1} - \binom{d+k-3}{d-1}.$ When $k\to\infty$, $\Nkd=\Theta(d^k)$ and when $d\to\infty$, $\Nkd=\Theta(k^d).$ Spherical harmonics with different degrees are orthogonal to each other, and their linear combinations can represent all square integrable functions over the sphere.

We use $\Pi_k$ to denote the projection operator to the degree-$k$ spherical harmonics space $\Ykd$. We use $\mathbb{Y}_{\le k,d}$ to denote the space of spherical harmonics up to degree $k$, and $\Pi_{\le k}\defeq \sum_{l=0}^{k}\Pi_l$ the projection operator to $\mathbb{Y}_{\le k,d}$. 

Spherical harmonics are closely related to Legendre polynomials. The degree-$k$ Legendre polynomial $\Pkd:\R\to\R$ is defined by the following recursive relationship
\begin{align}
	&P_{0,d}(t)=1,\quad P_{1,d}(t)=t,\\
	&P_{k,d}(t)=\frac{2k+d-4}{k+d-3}tP_{k-1,d}(t)-\frac{k-1}{k+d-3}P_{k-2,d}(t),\quad \forall k\ge 2.
\end{align}
Let $\Pkdb(t)\defeq \sqrt{\Nkd}\Pkd(t)$ be the normalized Legendre polynomial. Normalized Legendre polynomial is a set of complete orthonormal basis for square-integrable functions over $[-1,1]$ with respect to the measure $\mu_d(t)\defeq (1-t^2)^{\frac{d-3}{2}}\frac{\Gamma(d/2)}{\Gamma((d-1)/2)}\frac{1}{\sqrt{\pi}}$, which equals to the density of $x_1$ when $x=(x_1,\cdots,x_d)$ is uniformly drawn from sphere $\Sp$. In other words, $\dotp{\Pkdb}{\bar{P}_{k',d}}_{\mud}\defeq \int_{-1}^{1}\Pkdb(t)\bar{P}_{k',d}(t)\mu_d(t)\dd t=\ind{k=k'}.$

\paragraph{Properties of spherical harmonics and Legendre polynomials.} Our proof heavily replies on the following properties of spherical harmonics and Legendre polynomials.

Let $\{Y_{k,j}\}_{j=1}^{\Nkd}$ be an orthonormal basis of $\Ykd$. Then for any function $f:\Sp\to\R$ with $\|f\|_2<\infty$, there is a unique decomposition $f(\cdot)=\sum_{k\ge 0}\sum_{j=1}^{\Nkd}a_{k,j}Y_{k,j}(\cdot)$ with coefficients $\{a_{k,j}\}_{k\ge 0,1\le j\le \Nkd}$ that satisfies $\|f\|_2^2=\sum_{k\ge 0}\sum_{j=1}^{\Nkd}a_{k,j}^2$.

Spherical harmonics are the eigenfunctions of any inner-product kernels on the sphere, summarized by the following theorem \citep[Theorem 2.22]{atkinson2012spherical}.
\begin{theorem}[Funk-Hecke formula]\label{thm:funk-hecke}
	\sloppy Let $\sigma:[-1,1]\to\R$ be any one-dimensional function with $\int_{-1}^{1}|\sigma(t)|\mud(t)\dd t<\infty$, and $\lambda_k=\Nkd^{-1/2}\dotp{\sigma}{\Pkdb}_\mud$. Then for any function $Y_k\in \Ykd$,
	\begin{align}
		\textstyle{\forall x\in\Sp, \quad \E_{z\sim \Sp}[\sigma(x^\top z)Y_k(z)]=\lambda_k Y_k(x).}
	\end{align}
	In other words, $\Ykd$ is the space of eigenfunctions of the inner product kernel $K(x,z)\defeq \sigma(x^\top z)$ corresponding to the eigenvalue $\lambda_k.$
\end{theorem}

We can construct spherical harmonics using Legendre polynomials. For any degree $k\ge 0$, let $g_u:\Sp\to\R$ be the function $g_u(x)=\Pkdb(\dotp{x}{u})$. Then for any $u\in\Sp$, $g_u\in \Ykd$ and $\|g_u\|_2=1$.

In the worst-case, high-order spherical harmonics can be very spiky because their $L_\infty/L_2$ ratio is very large:
\begin{fact}\label{fact:infty-2-SH}
	For every fixed $k\ge 0, g\in \Ykd$ we have $\|g\|_\infty\le \sqrt{\Nkd}\|g\|_2$, and the equality is achieved by $g_u(\cdot)=\Pkdb(\dotp{\cdot}{u})$ for any $u\in\Sp$.
\end{fact}
	\section{Main Results}\label{sec:main-results}
In this section, we will first design a $L_\infty$-recovery algorithm that achieves polynomial sample complexity when the ground-truth function $f$ satisfies two conditions (Conditions~\ref{cond:decay} and \ref{cond:random}). We then establish these two conditions when $f$ is drawn from an isotropic Gaussian random fields (Lemma~\ref{lem:GRF}).

The first condition states that the spherical harmonics decomposition of the ground-truth $f$ decays at a proper rate. %
\begin{cond}\label{cond:decay}
	The ground-truth function $f$ satisfies $\|\Proj_k f\|_2\le c_1\Nkd^{-\alpha/2},\forall k\ge 0$ for some $c_1>0$ and $\alpha>0$. 
\end{cond}
We treat $\alpha$ as a constant that doesn't depend on the ambient dimension $d$. The parameter $\alpha>0$ is intuitively a notion of smoothness of the function $f$. This is because the derivatives of higher-degree spherical harmonics are larger. Hence, qualitatively speaking, functions with a faster decay (larger $\alpha$) is smoother.

Condition~\ref{cond:decay} holds for a wide range of functions. For example, any function of the form $g(\cdot)=h(\dotp{\cdot}{u})$, where $u\in\Sp$ and $h:[-1,1]\to \R$ with $\sup_{t\in[-1,1]} |h^{(k)}(t)|\le 1,\forall k\ge 0$, satisfies Condition~\ref{cond:decay} with parameter $\alpha=1$ and $c_1=1$ (Proposition~\ref{prop:smooth-inner-product-decay}). In addition, if two functions $g,h$ satisfy Condition~\ref{cond:decay}, so do their convex combinations $\theta g+(1-\theta)h,\forall \theta\in[0,1]$. Hence Condition~\ref{cond:decay} holds for \emph{any} two-layer NNs with bounded $L_1$ norm and infinitely smooth activation $h$ (e.g., exponential activation).

The following condition states that $f$ is not spiky when projected to the degree-$k$ spherical harmonics space. This condition is central to our analysis because it excludes the hard instances in the lower bounds (e.g., spiky functions constructed in Lemma~\ref{lem:relu-lb}).
\begin{cond}\label{cond:random}
The ground-truth function $f$ satisfies $\|\Proj_k f\|_\infty\le c_2\sqrt{\ln (k+1)}\|\Proj_k f\|_2$,$\forall k\ge 0$ for some $c_2>0$. 
\end{cond}

Condition~\ref{cond:random} requires that the $L_\infty$/$L_2$ ratio of $\Proj_k f$ is bounded by $c_2 \sqrt{\ln (k+1)}$, whereas the worst case ratio is $\sqrt{\Nkd}$ (Fact~\ref{fact:infty-2-SH}). As we will show later, Condition~\ref{cond:random} holds with high probability for random functions drawn from degree-$k$ spherical harmonics space $\Ykd$ (Lemma~\ref{lem:random-sh}) and functions drawn from isotropic Gaussian random fields (Lemma~\ref{lem:GRF}).

With Conditions~\ref{cond:decay} and \ref{cond:random}, our main theorem states that there exists an algorithm (described later in Alg.~\ref{alg:main}) that achieves $L_\infty$-recovery using only polynomial samples drawn from the uniform distribution over the sphere $\Sp.$
\begin{theorem}\label{thm:mainsc}
	Suppose the ground-truth function $f$ satisfies Conditions~\ref{cond:decay} and \ref{cond:random} for some fixed $\alpha\in (0,1]$ and $c_1,c_2>0$. If $d\ge 10\alpha^{-1}2^{5/\alpha}+2$, then for any $\epsilon>0,\delta>0$, with probability at least $1-\delta$ over the randomness of the data, Alg.~\ref{alg:main} outputs a function $g$ such that $\|f-g\|_\infty\le \epsilon$ using $O(\poly(c_1c_2,d,1/\epsilon,\ln 1/\delta)^{1/\alpha})$ samples.
\end{theorem}
In comparison, the classical kernel methods assume $f$ has a bounded RKHS norm, which is equivalent to assuming that $\|\Proj_k f\|_2$ decays at a rate determined by the choice of kernel. For any function $f$ with decay parameter $\alpha\in(0,1/2)$, Proposition~\ref{prop:rkhs-norm-of-f} shows that the RKHS norm of $f$ is infinite with respect to \emph{any} bounded inner product kernel (e.g., the RBF kernel), and thus violates the assumption of kernel methods. In contrast, Theorem~\ref{thm:mainsc} still implies polynomial sample complexity for $\alpha\in(0,1/2)$ thanks to the additional randomness condition (Condition~\ref{cond:random})

Our algorithm is stated in Alg.~\ref{alg:main}. On a high level, given any desired error $\epsilon>0$, the algorithm selects a truncation threshold $k\ge 0$ (Line~\ref{line:truncate}), and uses empirical risk minimization to find the best degree-$k$ polynomial approximation to the ground-truth $f$.

Instead of directly learning a degree-$k$ polynomial, we can also use two-layer neural networks with polynomial activation to approximate the function $f$. The algorithm and discussion are deferred to Appendix~\ref{app:nns}.
\begin{algorithm}[ht]
	\caption{$L_\infty$-learning via Low-degree Polynomial Approximation}
	\label{alg:main}
	\hspace*{\algorithmicindent} \textbf{Parameters:} $\alpha,c_1,c_2>0$, desired error $\epsilon>0$, and failure probability $\delta>0.$
	
	\hspace*{\algorithmicindent} \textbf{Input:} Dataset $\calD=\{(x_i,y_i)\}_{i=1}^{n}$ where $x_i\sim \Sp$ are i.i.d.~samples from the unit sphere, and $y_i=f(x_i)+\calN(0,1)$.
	\begin{algorithmic}[1]
		\State Set the truncation threshold $k\gets\inf_{l\ge 0}\{2c_1c_2(l+1)^{3/2}(N_{l+1,d})^{-\alpha/2}\le \epsilon/2\}.$\label{line:truncate}
		\State Define the function class
		\begin{align}
			\textstyle{\calF_k\gets\left\{g\in\mathbb{Y}_{\le k,d}: \|\Proj_l g\|_2\le c_1,\forall l\in[0,k]\right\}}.\label{equ:function-class}
		\end{align}
		\State Run empirical risk minimization:
		$
		g\gets \textstyle{\argmin_{h\in\calF_k} \sum_{i=1}^{n}(h(x_i)-y_i)^2.}
		$
		\State \textbf{Return} $g$.
	\end{algorithmic}
\end{algorithm}

We present a proof sketch of Theorem~\ref{thm:mainsc} in Section~\ref{sec:proofsketch-mainsc}, and defer the full proof to Appendix~\ref{app:pf-mainsc}.

\subsection{Instantiation of Theorem~\ref{thm:mainsc} on Gaussian Random Fields}
In this section, we instantiate Theorem~\ref{thm:mainsc} on isotropic Gaussian random fields.

Given any positive semi-definite covariance function $K:\Sp\times\Sp\to\R$, the mean-zero Gaussian random field is a collection of random variables $\{h(x)\}_{x\in \Sp}$ such that the distribution of any finite subset $(h(x_1),\cdots,h(x_n))$ is a Gaussian vector with covariance $\Sigma_{ij}=K(x_i, x_j).$ When the distribution is rotationally invariant, i.e., the distribution of $h(x_1),\cdots,h(x_n)$ equals to the distribution of $h(Rx_1),\cdots,h(Rx_n)$ for any rotation matrix $R\in\R^{d\times d}$, the covariance $K(x,x')$ only depends on the inner product $x^\top x'$ and can be written as $K(x,x')=\kappa(x^\top x')$ for some $\kappa:[-1,1]\to \R$. The corresponding GRF is called isotropic.

We focus on the case where the eigenvalues of the covariance (or equivalently, the Legendre polynomial decomposition of $\kappa$, by the Funk-Hecke formula) decays with a proper rate. Concretely, we assume $\kappa$ has the decomposition $\kappa(t)=\sum_{k\ge 0}\hat\kappa_k\Pkdb(t)$ where $\hat\kappa_k\le O(\Nkd^{-1/2-\alpha})$ for some $\alpha>0$. Later we will show that a function $f$ drawn from GRF with covariance $K(x,x')=\kappa(x^\top x')$ satisfies $\|\Proj_k f\|_2\le O(\Nkd^{-\alpha/2})$ and this inequality is tight. 
The decay rate $O(\Nkd^{-1/2-\alpha})$ is slightly faster than the decay of RBF kernels (given by $\kappa(t)=\exp(t)$), which is $\approx \Nkd^{-1/2}$ when $k$ is small \citep{minh2006mercer}.

The following theorem proves that Alg.~\ref{alg:main} can achieve $L_\infty$-recovery for function drawn from Gaussian random fields.
\begin{theorem}\label{thm:GRF-main}
	Let $f:\Sp\to \R$ be a function drawn from a Gaussian random field with covariance $K(x,x')=\kappa(x^\top x)$. Suppose for all $k\ge 0$, $\dotp{\kappa}{\Pkdb}_{\mud}\le c^2 \Nkd^{-1/2-\alpha}$ for some $c>0,\alpha>0$.
	Given any $\epsilon>0,\delta>0$, with probability at least $1-\delta$ over the randomness of $f$ and the dataset, Alg.~\ref{alg:main} outputs a function $g:\Sp\to\R$ such that $\|g-f\|_\infty\le \epsilon$ using $O(\poly(c,\epsilon^{-1},d,\ln 1/\delta)^{1/\alpha})$ samples.
\end{theorem}
To the best of our knowledge, Theorem~\ref{thm:GRF-main} is the first result that achieves a $L_\infty$-error guarantee for isotropic Gaussian processes using only polynomial samples drawn uniformly from the unit sphere. 

A closely related line of research to Theorem~\ref{thm:GRF-main} is the Gaussian process bandit problem, where the algorithm can adaptively query any data point and the goal is to maximize the function $f$ drawn from a Gaussian random field \citep{srinivas2010gaussian}. We can modify the GP-UCB algorithm in \citet{srinivas2010gaussian} to a $L_\infty$-recovery algorithm with \emph{adaptive} samples, and this modification, together with the analysis in \citet{vakili2021information}, lead to a polynomial sample complexity with the same condition as Theorem~\ref{thm:GRF-main}.\footnote{On a high level, at every iteration $t\ge 1$ the original GP-UCB algorithm selects the query $x_t$ that maximizes the upper confidence bound of $f$ \citep{srinivas2010gaussian}. 
\citet{srinivas2010gaussian} construct the upper confidence bound by analytically compute the posterior mean and variance of $f$ given any data points, assuming that the ground-truth $f$ is drawn from a Gaussian process prior. 
To get an algorithm for $L_\infty$-recovery, we can choose $x_t$ that maximizes the posterior variance of $f(\cdot)$. In this case, the analysis in \citep{srinivas2010gaussian} implies that with high probability after $n$ iterations, the $L_\infty$-error of the posterior mean is upper bounded by the maximum information gain, denoted by $\sqrt{\gamma_n/n}$. 
Combining with the refined analysis in \citet{vakili2021information}, we can upper bound the information gain $\gamma_n$ using the spectrum decay of $\kappa$, which leads to a polynomial sample complexity in our setting.
}
In comparison, our algorithm only requires samples from the uniform distribution while GP-UCB must be adaptive. In addition, Theorem~\ref{thm:mainsc} holds for general functions with Condition~\ref{cond:decay} and \ref{cond:random} while the analysis of \citet{srinivas2010gaussian} is specialized to Gaussian processes.

We prove Theorem~\ref{thm:GRF-main} by establishing Conditions~\ref{cond:decay} and \ref{cond:random} for functions drawn from isotropic Gaussian random fields using the following lemma, and then directly invoking Theorem~\ref{thm:mainsc}.
\begin{lemma}\label{lem:GRF}
	In the setting of Theorem~\ref{thm:GRF-main}, with probability at least $1-\delta$ we have
	\begin{align}\label{equ:GRF-1}
		\forall k\ge 0,&\quad \|\Proj_k f\|_\infty\le 5\sqrt{2\ln(6/\delta)+2(d^2+1)\ln(k+1)}\|\Proj_k f\|_2,
	\end{align}
	and
	\begin{align}
		\forall k\ge 0,&\quad \|\Proj_k f\|_2\le 3c\sqrt{\ln(2/\delta)}\Nkd^{-\alpha/2}.\label{equ:GRF-2}
	\end{align}
\end{lemma}

We the proof of Lemma~\ref{lem:GRF} is deferred to Section~\ref{sec:proofsketch-GRF}.

\subsection{Proof Sketch of Theorem~\ref{thm:mainsc}}\label{sec:proofsketch-mainsc}
In this section, we present the proof sketches of Theorem~\ref{thm:mainsc}. On a high level, we prove that (a) the ground-truth $f$ can be approximated by a low-degree polynomial with a small $L_\infty$-error, and (b) learning a low-degree polynomial in $L_\infty$-error only requires polynomial samples.

\begin{proof}[Proof sketch of Theorem~\ref{thm:mainsc}]
	For better exposition, in the following we present the proof sketch for the case $\alpha=1/2$, and the general case is proved similarly.
	
	For any fixed threshold $k\ge 0$, we first upper bound the $L_\infty$-distance between the ground-truth $f$ and its low-degree components $\Proj_{\le k}f$. Concretely, 
	\begin{align}
		\textstyle{\|f-\Proj_{\le k}f\|_\infty=\|\sum_{l>k}\Proj_l f\|_\infty\le \sum_{l>k}\|\Proj_l f\|_\infty.}
	\end{align}
	Under Conditions~\ref{cond:decay} and \ref{cond:random}, the term $\|\Proj_l f\|_\infty$ decays at rate $\sqrt{\ln (l+1)}\Nld^{-1/2}$. Since $\Nld^{-1/2}\approx \min\{l^d,d^l\}^{-1/2}$ decays very fast, we get
	\begin{align}\label{equ:pf-thm-mainsc-1}
		\textstyle{\|f-\Proj_{\le k}f\|_\infty\le \sum_{l>k}\sqrt{\ln (l+1)}\Nld^{-1/2}\lesssim \sqrt{\ln (k+1)}\Nkd^{-1/2}}.
	\end{align}

	Next we show that the low-degree components $\Proj_{\le k}f$ can be learned w.r.t. $L_\infty$-error using polynomial samples because the $L_\infty$-error of a low-degree polynomial is upper bounded by its $L_2$-error. Indeed, Fact~\ref{fact:infty-2-SH} states that $\|h\|_\infty\le \sqrt{\Nkd}\|h\|_2,\forall h\in\Ykd$. Then for any low-degree polynomial $g\in\mathbb{Y}_{\le k,d}$,
	\begin{align}\label{equ:pf-thm-mainsc-2}
		\textstyle{\|g-\Proj_{\le k}f\|_\infty=\|\Proj_{\le k}(g-f)\|_\infty\le \sum_{l=0}^{k} \|\Proj_{l}(g-f)\|_\infty \le \sum_{l=0}^{k} \|\Proj_{l}(g-f)\|_2 \Nld^{1/2}.}
	\end{align}
	When $g\in\mathbb{Y}_{\le k,d}$, we have $\|g-\Proj_{\le k}f\|_2^2=\|\Proj_{\le k}(g-f)\|_2^2=\sum_{l=0}^{k}\|\Proj_{l}(g-f)\|_2^2$. Continuing Eq.~\eqref{equ:pf-thm-mainsc-2} by applying Cauchy-Schwarz, we get
	\begin{align}\label{equ:pf-thm-mainsc-3}
		\textstyle{\|g-\Proj_{\le k}f\|_\infty\le \Nkd^{1/2}\sqrt{k+1}\|\Proj_{\le k}(g-f)\|_2=\Nkd^{1/2}\sqrt{k+1}\|g-\Proj_{\le k}f\|_2.}
	\end{align}
	Now we can choose an threshold $k\ge 0$ to balance the two terms in Eq.~\eqref{equ:pf-thm-mainsc-1} and Eq.~\eqref{equ:pf-thm-mainsc-3}. For any desired error level $\epsilon>0$, we can choose an $k$ such that $\sqrt{\ln (k+1)}\Nkd^{-1/2}=\Theta(\epsilon/2)$ and get  
	\begin{align}\label{equ:pf-thm-mainsc-4}
		\textstyle{\|g-f\|_\infty\le \|g-\Proj_{\le k}f\|_\infty+\|f-\Proj_{\le k}f\|_\infty\lesssim \poly(1/\epsilon)\|g-\Proj_{\le k}f\|_2+\epsilon/2.}
	\end{align}

	Finally, for any truncation threshold $k>0$, $\Proj_{\le k}f$ is a low-degree polynomial and belongs to the family $\calF_k$ defined in Eq.~\eqref{equ:function-class}. Therefore classic statistical learning theory implies that empirical risk minimization outputs a function $g$ with $\|g-\Proj_{\le k}f\|_2\le \poly(\epsilon)$ using only $\poly(1/\epsilon)$ samples (Lemma~\ref{lem:ERM-l2-kernel}), which completes the proof.
\end{proof}

\subsection{Proof of Lemma~\ref{lem:GRF}}\label{sec:proofsketch-GRF}
To prove Lemma~\ref{lem:GRF}, we first characterize an isotropic Gaussian random field in the spherical harmonics expansion.

Let $f:\Sp\to\R$ be a function drawn from an isotropic Gaussian random field with covariance $\kappa:[0,1]\to\R$, and $\{Y_{k,j}\}_{k\ge 0,1\le j\le \Nkd}$ a set of orthonormal spherical harmonics basis. We will show that the projection of $f$ to the degree-$k$ spherical harmonics space is isotropic. In other words, $\{\dotp{f}{Y_{k,j}}\}_{1\le j\le \Nkd}$ are i.i.d. random variables.

Indeed, by \citet[Theorem 5.5]{lang2015isotropic}, $f$ admits the following spherical harmonics decomposition
\begin{align}\label{equ:pfs-GRF-1}
	\textstyle{f(\cdot)\stackrel{d}{=}\sum_{k\ge 0}\(\hat\kappa_k^{1/2}\Nkd^{-1/4}\sum_{j=1}^{\Nkd}a_{k,j}Y_{k,j}(\cdot)\),}
\end{align}
where $a_{k,j}$ are i.i.d.~unit Gaussian random variables. Hence, to prove Lemma~\ref{lem:GRF} we only need to examine the property of a random function drawn from the spherical harmonics space $\Ykd$, which is a $\Nkd$-dimensional Hilbert space.

The following lemma shows that a random spherical harmonics is not spiky because its $L_\infty$/$L_2$ ratio is upperbounded by $O(d\sqrt{\ln k})$ with high probability, whereas the worst case ratio is $\sqrt{\Nkd}=\Omega(\min\{d^{k/2},k^{d/2}\}).$
\begin{lemma}\label{lem:random-sh}
	For any fixed $k\ge 0$, let $\{Y_{k,j}\}_{j=1}^{\Nkd}$ be any set of orthonormal basis for degree-$k$ spherical harmonics $\Ykd$.
	Let $g=\sum_{j=1}^{\Nkd}a_{j}Y_{k,j}$ be a random spherical harmonics where $a_j\sim \calN(0,1)$ are independent unit Gaussian random variables. For any $\delta>0$ we have, with probability at least $1-\delta$,
	\begin{align}\label{equ:random-sh}
		\|g\|_\infty\le 5\sqrt{\ln(3/\delta)+2d^2\ln(k+1)}\|g\|_2.
	\end{align}
\end{lemma}
Lemma~\ref{lem:random-sh} is a high-dimensional version of \citet[Theorem 2]{burq2014probabilistic}. The proof of \citet{burq2014probabilistic} relies on the Sobolev embedding theorem, which treats the dimension $d$ as a constant. In contrast, we compute the exact dependency on the dimension $d$ by instantiating the Riesz representation theorem on the space of spherical harmonics and then applying a uniform convergence argument. 

In the following, we present a proof sketch of Lemma~\ref{lem:random-sh}. The full proof is deferred to Appendix~\ref{app:random-sh}.

\begin{proof}[Proof Sketch of Lemma~\ref{lem:random-sh}]
	To prove the $L_\infty$/$L_2$ norm ratio of $g$, we first invoke Lemma~\ref{lem:riesz-SH} which states that
	\begin{align}
		\forall x\in\Sp,\quad g(x)=\sqrt{\Nkd}\dotp{g}{\Pkdb(\dotp{x}{\cdot})}.
	\end{align}
	Since $\Pkdb(\dotp{x}{\cdot})\in \Ykd$, Lemma~\ref{lem:riesz-SH} is an instantiation of the Riesz representation theorem on the space $\Ykd$. The Riesz representation theorem states that for a Hilbert space, every continuous linear functional (in this case, the evaluation functional ${\rm{ev}}_x:g\to g(x)$) can be represented by the inner product with an element in the space (in this case, $\sqrt{\Nkd}\Pkdb(\dotp{x}{\cdot})$).
	
	For any fixed $x\in\Sp$, because $g=\Proj_k f$ is a Gaussian vector in the $\Nkd$-dimensional space $\Ykd$ and $\Pkdb(\dotp{x}{\cdot})\in\Ykd$ is a fixed vector, the function value $g(x)$ has a Gaussian distribution.
	Formally speaking, we can write $\Pkdb(\dotp{x}{\cdot})=\sum_{j=1}^{\Nkd}u_{k,j}Y_{k,j}(\cdot)$ for some fixed parameters $u_{k,j}.$ Let $\vec{a}_k=[a_{k,j}]_{1\le j\le \Nkd}$ and $\vec{u}_k=[u_{k,j}]_{1\le j\le \Nkd}$, then we get
	\begin{align}
		g(x)=\sqrt{\Nkd}\dotp{g}{\Pkdb(\dotp{x}{\cdot})}=\sqrt{\Nkd}\dotp{\vec{a}_k}{\vec{u}_k}\sim \sqrt{\Nkd}\calN(0,\|\vec{u}_k\|_2^2).
	\end{align}
	Since $\|\vec{u}_k\|_2=\|\Pkdb(\dotp{x}{\cdot})\|_2=1$ and $\|g\|_2=\|\vec{a}_k\|_2\approx \sqrt{\Nkd}$, by concentration inequality of Gaussian vectors (Lemma~\ref{lem:gaussian-proj-concentration}) we get for any fixed $x\in\Sp$, with high probability
	$
	|g(x)|\lesssim \|\vec{a}_k\|_2=\|g\|_2.
	$
	Finally, we can use a covering number argument to prove a uniform convergence of all $x\in\Sp$. Hence, we prove that with high probability, $\forall x\in\Sp, |g(x)|\le \tilde{O}(d\sqrt{\ln k})\|g\|_2$, which implies Eq.~\eqref{equ:random-sh}.
\end{proof}

With Lemma~\ref{lem:random-sh}, we can now prove Lemma~\ref{lem:GRF}.
\begin{proof}[Proof of Lemma~\ref{lem:GRF}]
	Recall that \citet[Theorem 5.5]{lang2015isotropic} gives the following spherical harmonics decomposition
	\begin{align}
		\textstyle{f(x)\stackrel{d}{=}\sum_{k\ge 0}\(\hat\kappa_k^{1/2}\Nkd^{-1/4}\sum_{j=1}^{\Nkd}a_{k,j}Y_{k,j}(x)\)}
	\end{align}
	where $a_{k,j}\sim \calN(0,1)$ are independent Gaussian random variables. By Lemma~\ref{lem:random-sh}, for any fixed $k\ge 0$, with probability at least $1-\delta/(2(k+1)^2)$ we have
	\begin{align}
		\|\Proj_k f\|_\infty&\le 5\sqrt{\ln(6(k+1)^2/\delta)+2d^2\ln(k+1)}\|\Proj_k f\|_2\\
		&\le 5\sqrt{2\ln(6/\delta)+2(d^2+1)\ln(k+1)}\|\Proj_k f\|_2.
	\end{align}
	By union bound over $k$, with probability at least $1-\delta$ we get
	\begin{align}
		\forall k\ge 0, \quad \|\Proj_k f\|_\infty\le 5\sqrt{2\ln(6/\delta)+2(d^2+1)\ln(k+1)}\|\Proj_k f\|_2,
	\end{align}
	which proves Eq.~\eqref{equ:GRF-1}.
	
	Now we prove the second part of lemma. Since $\{Y_{k,j}\}_{j=1}^{\Nkd}$ forms an orthonormal basis of $\Ykd$, we get
	\begin{align}\label{equ:pf-GRF-1}
		\textstyle{\|\Proj_k f\|_2^2=\hat{\kappa}_k\Nkd^{-1/2}\sum_{j=1}^{\Nkd}a_{k,j}^2\le c^2\Nkd^{-1-\alpha}\sum_{j=1}^{\Nkd}a_{k,j}^2.}
	\end{align}
	For any fixed $k\ge 0$, since $a_{k,j}$ are i.i.d. unit Gaussian random variables, by the concentration of the norm of Gaussian vectors \citep[Lemma 1]{laurent2000adaptive}, we have
	\begin{align}
		\textstyle{\forall t>0, \quad \Pr\(\sum_{j=1}^{\Nkd}a_{k,j}^2\ge \Nkd+2\sqrt{\Nkd}\sqrt{t}+2t\)\le \exp(-t).}
	\end{align}
	
	Take $t=\ln (2(k+1)^2/\delta)$. Note that $\Nkd\ge k\ge \ln(k+1)$. As a result, for all $k\ge 0$ we get
	\begin{align}
		\Nkd+2\sqrt{\Nkd}\sqrt{t}+2t\le 9\Nkd\ln(2/\delta).
	\end{align}
	Consequently,
	\begin{align}
		\textstyle{\Pr\(\sum_{j=1}^{\Nkd}a_{k,j}^2\ge 9\Nkd \ln(2/\delta)\)\le (k+1)^{-2}\delta/2.}
	\end{align}
	Combining with Eq.~\eqref{equ:pf-GRF-1} and union bound over $k$, with probability at least $1-\delta$ we get
	\begin{align}
		\textstyle{\forall k\ge 0,\quad \|\Proj_k f\|_2\le c\Nkd^{-1/2-\alpha/2}\(\sum_{j=1}^{\Nkd}a_{k,j}^2\)^{1/2} \le 3c\sqrt{\ln(2/\delta)}\Nkd^{-\alpha/2},}
	\end{align}
	which proves Eq.~\eqref{equ:GRF-2}.
\end{proof}

	\section{Lower Bounds}\label{sec:lowerbounds}
In this section, we present two lower bounds to motivate our Condition~\ref{cond:random}. Both lower bounds hold for any algorithm that can \emph{adaptively} choose its data point $x_i$ and observes a noisy signal $f(x_i)+\calN(0,1)$, where $f$ denotes the ground-truth function. Our lower bounds may be of independent interest.

\paragraph{Lower bounds for functions with decay rate $\Nkd^{-1/2}$.}
The following lemma proves that, in the worst case, $L_\infty$-recovery is hard even when the function's spherical harmonics decomposition decays at a rate of $\Nkd^{-1/2}.$
\begin{lemma}\label{lem:relu-lb}
	For a fixed integer $k\ge 4$ and $\beta_k\in (0,1)$, define $\calF_k=\{\beta_k\Pkd(\dotp{\cdot}{u}):u\in\Sp\}$ be the hypothesis class. 
	For any fixed algorithm, let $E_{f,n}$ be the probability that the algorithm outputs $\hat{f}$ such that $\|\hat{f}-f\|_\infty\le \beta_k/4$ using $n$ samples when the ground-truth function is $f$. Then if $n<\Nkd\beta_k^{-2}$,
	$\min_{f\in \calF_k}E_{f,n}\le 1/2.$
\end{lemma}
Since $\|\Pkd(\dotp{\cdot}{u})\|_2=\Nkd^{-1/2}$, the function class $\calF_k$ (when $\beta_k=1$) is a subset of functions that satisfies Condition~\ref{cond:decay} with $\alpha=1$. Therefore, no algorithm can achieve polynomial sample complexity for $L_\infty$-recovery with only the smoothness condition (Condition~\ref{cond:decay}).

Lemma~\ref{lem:relu-lb} is proved by showing that no algorithm can distinguish all the functions $f\in\calF_k$ using $o(\Nkd)$ samples because the average signal-to-noise ratio of any data point is roughly $\Nkd^{-1/2}$. Hence, the worst-case sample complexity is at least $\Omega(\Nkd)$. The proof is deferred to Appendix~\ref{app:pf-lem-relu-lb}.

\paragraph{Lower bounds for two-layer ReLU neural networks.}
We first formally define the class of two-layer neural networks used in this paper. Let $\nnrelu(L_p)$ be the family of two layer neural networks (NNs) with $L_p$-norm bounds. Formally speaking, 
\begin{align}
	\nnrelu(L_p)=\{g(x)\defeq\E_{\xi\sim \Sp}[\sigma(x^\top \xi)c(\xi)]:\|c\|_p\le 1\},
\end{align}
where $\sigma$ is the ReLU activation and $c:\Sp\to\R$ is the weight of the NN. Classical finite width neural networks belong to $\nnrelu(L_1)$ because their weights $c$ can be represented by the mixtures of Dirac measures.

The following theorem shows that learning two-layer neural networks with ReLU activation is statistically hard even when the NN has a constant norm. The lower bound holds for $\nnrelu(L_2)$, which is a subset of $\nnrelu(L_1)$.
\begin{theorem}\label{thm:relu-lb}
	Given the hypothesis class $\nnrelu(L_2)$. If an algorithm, when running on every possible instance $f\in\nnrelu(L_2)$, takes in $n$ data points uniformly sampled from the sphere $\Sp$ and outputs a function $g$ such that $\|f-g\|_\infty\le \epsilon$ with probability at least $1/2$, then $n\ge \Omega\(\(0.002\epsilon^{-1}d^{-7/4}\)^{d/2}\)$. As a corollary, the minimax sample complexity of learning $\nnrelu(L_2)$ with $L_\infty$-error $\epsilon=O(d^{-7/4})$ requires at least $2^{d}$ samples.
\end{theorem}
Theorem~\ref{thm:relu-lb} does not contradict with existing results on the recovery of two-layer neural networks \citep{zhong2017recovery,zhou2021local} because they focus on the finite-width case while our lower bound holds for infinite-width neural networks. Compared with the lower bound in \citet{dong2021provable}, Theorem~\ref{thm:relu-lb} does not rely on the bias term in the ReLU activation to kill the signal. Instead, we invoke the Funk-Hecke formula (Theorem~\ref{thm:funk-hecke}) to show that two-layer ReLU NNs can represent spiky functions with constant norm.

Theorem~\ref{thm:relu-lb} is proved by showing that $\calF_k$ defined in Lemma~\ref{lem:relu-lb} is a subset of $\nnrelu(L_2)$ if we take $\beta_k\approx k^{-2}$. The proof is deferred to Appendix~\ref{app:pf-thm-relu-lb}.
	\section{Conclusion}\label{sec:conclusion}
In this paper, we make some initial steps toward $L_\infty$-recovery for nonlinear models by proving a polynomial sample complexity bound for random function drawn from Gaussian random fields. We also prove a $\exp(d)$ sample complexity lower bound for recovering the worst-case infinite-width two-layer neural nets with unbiased ReLU activation, which may be of independent interest.

For future works, we raise the following open questions:
\begin{enumerate}
	\item To instantiate Condition~\ref{cond:random}, this paper focuses on functions $f$ drawn from Gaussian random fields because they have \emph{independent} components in the spherical harmonics space. However, Condition~\ref{cond:random} also holds when $f$ has correlated components. For example, when $\Proj_k f=\sum_{j=1}^{\Nkd}a_{k,j}Y_{k,j}$ where $[a_{k,j}]_{j=1}^{\Nkd}$ lies on the $(\Nkd-1)$-dimensional sphere. Is it possible to prove Condition~\ref{cond:random} for functions drawn from other distribution?
	\item A two-layer single-neuron neural nets with exponential activation, i.e., functions of the form $g(\cdot)=\exp(\dotp{\cdot}{u})$ for some $u\in\Sp$, does not satisfy Condition~\ref{cond:random}. In fact, $\Proj_k g$ is the most spiky function in $\Ykd$ because $\Proj_k g=\lambda_k \Pkdb(\dotp{\cdot}{u})$. Can we find a natural (random) subset of two-layer neural networks that satisfy Condition~\ref{cond:random}?
\end{enumerate}
	
	\subsection*{Acknowledgment}
	The authors would like to thank Ruixiang Zhang, Yakun Xi, Yuhao Zhou, Jason D. Lee for helpful discussions. The authors would also like to thank anonymous reviewers for the references to additional related works. The authors would like to thank the support of NSF CIF 2212263.
	
	\bibliographystyle{plainnat}
	\bibliography{all.bib}
	\newpage
	\appendix
	\section*{List of Appendices}
	\startcontents[sections]
	\printcontents[sections]{l}{1}{\setcounter{tocdepth}{2}}
	\newpage
	
	\section{Missing Proofs}

\subsection{Proof of Theorem~\ref{thm:mainsc}}\label{app:pf-mainsc}
In the following, we first state two lemmas that are critical to the proof of Theorem~\ref{thm:mainsc}.

The next lemma proves that the empirical risk minimization step used in Alg.~\ref{alg:main} outputs a function with small $L_2$ loss, whose proof is deferred to Appendix~\ref{app:pf-ERM-l2-kernel}.
\begin{lemma}\label{lem:ERM-l2-kernel}
	Suppose the function $f:\Sp\to \R$ satisfies Condition~\ref{cond:decay} for some fixed $\alpha\in (0,1],c_1,c_2>0$. For any $\epsilon>0$, let $k=\inf_{l\ge 0}\{2c_1c_2(l+1)^{3/2}(N_{l+1,d})^{-\alpha/2}\le \epsilon/2\}$.
	
	Let $\calF_k\gets\left\{g\in\mathbb{Y}_{\le k,d}: \|\Proj_l g\|_2\le c_1,\forall l\in[0,k]\right\}$ be the function class defined in Alg.~\ref{alg:main}.
	For a given dataset $\{(x_i,y_i)\}_{i=1}^{n}$, let $\hat\calL(h)\defeq \frac{1}{n}\sum_{i=1}^{n}(h(x_i)-y_i)^2$ be the empirical $L_2$ loss, and $g=\argmin_{h\in\calF_k}\hat\calL(h).$
	
	For any $\delta>0,\epsilon_1>0$, when $d\ge \max\{2e,4/\alpha\}$ and the number of samples $n\ge \Omega(\poly(c_1c_2,1/\epsilon)^{1/\alpha}\poly(1/\epsilon_1,\ln(1/\delta)))$, with probability at least $1-\delta$,
	\begin{align}
		\|\Proj_{\le k} (f-g)\|_2=\|\Proj_{\le k} f-g\|_2\le \epsilon_1.
	\end{align}
\end{lemma}

The following lemma proves that with Conditions~\ref{cond:decay} and \ref{cond:random}, $\|f-\Proj_{\le k} g\|_\infty$ can be upper bounded by $\|\Proj_{\le k}(f-g)\|_2$ for properly chosen $k$.
\begin{lemma}\label{lem:main}
	Suppose the function $f:\Sp\to \R$ satisfies Conditions~\ref{cond:decay} and \ref{cond:random} for some fixed $\alpha\in (0,1],c_1,c_2>0$, and $d\ge 10\alpha^{-1}2^{5/\alpha}+2$. For any $\epsilon>0$, define $k=\inf_{l\ge 0}\{2c_1c_2(l+1)^{3/2}(N_{l+1,d})^{-\alpha/2}\le \epsilon/2\}.$
	Then for any function $g:\Sp\to \R$ with $\|\Proj_{\le k} (f-g)\|_2\le \frac{1}{4}\epsilon^{3/\alpha+1}(4c_1c_2)^{-3/\alpha}d^{-4/\alpha}$, we have
	$
	\|f-\Proj_{\le k} g\|_\infty\le \epsilon.
	$
\end{lemma}
Proof of Lemma~\ref{lem:main} is deferred to Appendix~\ref{app:pf-lem-main}.

Now we are ready to prove Theorem~\ref{thm:mainsc}.
\begin{proof}[Proof of Theorem~\ref{thm:mainsc}]
	Let $\epsilon_1=\frac{1}{4}\epsilon^{3/\alpha+1}(4c_1c_2)^{-3/\alpha}d^{-4/\alpha}$. We prove Theorem~\ref{thm:mainsc} in the following two steps.
	\paragraph{Step 1: upper bound the population $L_2$ loss.} In this step, we use classic statistical learning tools to show that the ERM step (i.e., $g=\argmin_{h\in\calF_k} \sum_{i=1}^{n}(h(x_i)-y_i)^2$) returns a function $g$ with small $L_2$ loss. In particular, by Lemma~\ref{lem:ERM-l2-kernel} we get
	$
	\|\Proj_{\le k} (f-g)\|_2\le \epsilon_1.
	$
	
	\paragraph{Step 2: upper bound the $L_\infty$-error via truncation.} In this step we show that with Conditions~\ref{cond:decay} and \ref{cond:random} on the ground-truth function, any function $g$ with a small $L_2$-error will also have a small $L_\infty$-error when projected to the low-degree spherical harmonics space. Formally speaking, invoking Lemma~\ref{lem:main} we get
	$
	\|\Proj_{\le k} (f-g)\|_2\le \epsilon_1 \implies \|f-\Proj_k g\|_\infty\le \epsilon.
	$
	
	Finally, since $g\in\calF_k\subset \mathbb{Y}_{\le k,d}$, we get $g=\Proj_k g$. Hence, combining these two steps we prove the desired result.
\end{proof}

\subsection{Proof of Lemma~\ref{lem:ERM-l2-kernel}}\label{app:pf-ERM-l2-kernel}
In the following we prove Lemma~\ref{lem:ERM-l2-kernel}.
\begin{proof}[Proof of Lemma~\ref{lem:ERM-l2-kernel}]
	We prove Lemma~\ref{lem:ERM-l2-kernel} in two steps.
	
	\paragraph{Step 1: expressivity.} In this step, we prove that $\Proj_{\le k}f\in\calF_k.$ Indeed, by Condition~\ref{cond:decay} we get
	\begin{align}
		\|\Proj_k f\|_2\le c_1\Nkd^{-\alpha/2}\le c_1,
	\end{align}
	meaning that $\Proj_{\le k}f\in\calF_k.$
	
	Consequently, using the definition $g=\argmin_{h\in\calF_k} \hat{\calL}(g)$ we have $\hat{\calL}(g)\le \hat{\calL}(\Proj_{\le k} f).$
	
	\paragraph{Step 2: uniform convergence.} In this step, we prove that using $$n=\Omega(\poly(c_1,\Nkd,\ln(1/\delta),1/\epsilon_1))$$ samples, Alg.~\ref{alg:main} outputs a function $g\in\calF$ such that 
	\begin{align}\label{equ:ERM-l2-kernel-1}
		\|g-\Proj_{\le k}f\|_2\le \epsilon_1.
	\end{align}
	
	To this end, 
	by the uniform convergence of $\calF_k$ (Lemma~\ref{lem:uniform-convergence-kernel}), when $$n= \Omega(\poly(c_1,\Nkd,\ln(1/\delta),1/\epsilon_1)),$$ with probability at least $1-\delta$,
	\begin{align}\label{equ:ERM-l2-kernel-0}
		&\|g-f\|_2^2\le \hat\calL(g)+\epsilon_1^2/2\le \hat\calL(\Proj_{\le k}f)+\epsilon_1^2/2\le\|\Proj_{\le k}f-f\|_2^2+\epsilon_1^2.
	\end{align}
	Since $\calF_k\subseteq \mathbb{Y}_{\le k,d}$, by the Parseval's identity we get
	\begin{align}
		\forall h\in\calF_k, \quad \|h-f\|_2^2&=\|\Proj_{\le k}(h-f)\|_2^2+\|\Proj_{>k}(h-f)\|_2^2\\
		&=\|\Proj_{\le k}(h-f)\|_2^2+\|\Proj_{>k}f\|_2^2=\|h-\Proj_{\le k}f\|_2^2+\|f-\Proj_{\le k}f\|_2^2.
	\end{align}
	Note that $g\in\calF_k$ and $\Proj_{\le k}f\in\calF_k$. Combining with Eq.~\eqref{equ:ERM-l2-kernel-0} we get
	\begin{align}
		\|g-\Proj_{\le k} f\|_2^2\le \epsilon_1^2.
	\end{align}
	Finally, by the choice of $k$ and Proposition~\ref{prop:truncation-upperbound}, $\Nkd=\poly(c_1c_2,1/\epsilon)^{1/\alpha}$, which means that 
	\begin{align}
		n= \Omega(\poly(c_1,\Nkd,\ln(1/\delta),1/\epsilon_1))=\Omega(\poly(c_1c_2,1/\epsilon)^{1/\alpha}\poly(\ln(1/\delta),1/\epsilon_1)).
	\end{align}
\end{proof}

The following lemma proves uniform convergence results for the function class $\calF_k.$
\begin{lemma}\label{lem:uniform-convergence-kernel}
	In the setting of Lemma~\ref{lem:ERM-l2-kernel}, for any $\delta>0,\epsilon_1>0$ and $n\ge \Omega(\poly(c_1,\Nkd,\ln(1/\delta),1/\epsilon_1)),$ with probability at least $1-\delta$ we have
	\begin{align}
		\sup_{g\in\calF_k}|\|g-f\|_2^2-\hat\calL(g)|\le \epsilon_1.
	\end{align}
\end{lemma}
\begin{proof}
	We prove this lemma using the Rademacher complexity of kernel methods \citep{bartlett2002rademacher}. First we upper bound the Rademacher complexity of $\calF_k$. Let $x_1,\cdots,x_n$ be a set of data points and $\hat{R}_n(\calF_k)$ the empirical Rademacher complexity of $\calF_k$, defined by
	\begin{align}
		\hat{R}_n(\calF_k)=\frac{1}{n}\E_{\sigma_1,\cdots,\sigma_n\sim \{-1,1\}^n}\[\sup_{g\in \calF_k}\abs{\sum_{i=1}^{n}\sigma_i g(x_i)}\].
	\end{align}
	Recall that $\{Y_{k,j}\}_{j=1}^{\Nkd}$ is an orthonormal basis of $\Ykd$, and any function $g\in\calF_k$ can be written as $g(x)=\sum_{l=0}^{k}\sum_{j=1}^{\Nld}a_{l,j}Y_{l,j}(x)$ where $\sum_{j=1}^{\Nkd}a_{l,j}^2\le c_1^2,\forall l\in[0,k]$. Hence, after defining $\phi_k(x)\defeq [Y_{l,j}(x)]_{l\in[0,k],j\in[\Nld]}$ as the feature vector, and $\vec{a}\defeq [a_{l,j}]_{l\in[0,k],j\in[\Nld]}$, we have $g(x)=\dotp{\phi_k(x)}{\vec{a}}$ and $\|\vec{a}\|_2\le \sqrt{k+1}c_1$.
	
	Let $k(x,x')=\dotp{\phi_k(x)}{\phi_k(x')}$ be the kernel function. Then by the fact that $\sum_{j=1}^{\Nld}Y_{l,j}(x)^2=\Nld,\forall l\ge 0$ \citep[Theorem 2.9]{atkinson2012spherical}, we have
	\begin{align}
		k(x,x)=\sum_{l=0}^{k}\sum_{j=1}^{\Nld}Y_{l,j}(x)^2=\sum_{l=0}^{k}\Nkd\le (k+1)\Nkd.
	\end{align}
	By \citet[Lemma 22]{bartlett2002rademacher} we get $\hat{R}_n(\calF_k)\le \frac{2(k+1)\sqrt{\Nkd}}{\sqrt{n}}.$
	
	Since for any $x$, we get $g(x)\le \|\phi_k(x)\|_2\|a\|_2=c_1(k+1)\sqrt{\Nkd}$, the $L_2$ loss is $(2c_1(k+1)\sqrt{\Nkd})$-Lipschitz. As a result, \citet[Theorem 3]{kakade2008complexity} implies that with probability at least $1-\delta$, $\forall g\in\calF_k$
	\begin{align}
		|\|g-f\|_2^2-\hat\calL(g)|=|\E[\hat\calL(g)]-\hat\calL(g)|\lesssim \frac{c_1(k+1)^2\Nkd}{\sqrt{n}}+c_1(k+1)^2\Nkd\sqrt{\frac{\ln(1/\delta)}{n}}.
	\end{align}
	Note that $\Nkd\ge k$. As a result, when $n\ge \Omega(\poly(c_1,\Nkd,\ln(1/\delta),1/\epsilon_1))$, we get 
	\begin{align}
		\forall g\in\calF_k,\quad |\|g-f\|_2^2-\hat\calL(g)|\le \epsilon_1.
	\end{align}
	which proves the desired result.
\end{proof}

\subsection{Proof of Lemma~\ref{lem:truncation}}
In the following we present and prove Lemma~\ref{lem:truncation}, which is used to prove Lemma~\ref{lem:main}.
\begin{lemma}\label{lem:truncation}
	Suppose the function $f:\Sp\to \R$ satisfies Conditions~\ref{cond:decay} and \ref{cond:random} for some fixed $\alpha\in (0,1]$ and $c_1,c_2>0$.
	When $d\ge 10\alpha^{-1}2^{5/\alpha}+2$, we have
	\begin{align}
		\|f-\Proj_{\le k-1}f\|_\infty\le 2c_1c_2k^{3/2}(\Nkd)^{-\alpha/2},\quad\forall k\ge 1.
	\end{align}
\end{lemma}
\begin{proof}[Proof of Lemma~\ref{lem:truncation}]
	Let $c=c_1c_2$. By basic algebra we get
	\begin{align}
		\|f-\Proj_{\le k-1}f\|_\infty=\norm{\sum_{l\ge 0}\Proj_{l}f-\Proj_{\le k-1}f}_\infty=\norm{\sum_{l\ge k}\Proj_{l}f}_\infty\le \sum_{l\ge k}c\sqrt{\ln (l+1)}(\Nld)^{-\alpha/2}.
	\end{align}
	Therefore we only need to prove 
	\begin{align}\label{equ:truncation-1}
		\sum_{l\ge k}c\sqrt{\ln (l+1)}(\Nld)^{-\alpha/2}\le 2ck^{3/2}(\Nkd)^{-\alpha/2}.
	\end{align}
	Recall that $\Nld=\frac{2l+d-2}{l+d-2}\frac{\Gamma(l+d-1)}{\Gamma(l+1)\Gamma(d-1)}$. It follows that 
	\begin{align}\label{equ:truncation-4}
		&\sum_{l\ge k}c\sqrt{\ln (l+1)}(\Nld)^{-\alpha/2}\le \;\sum_{l\ge k}c\sqrt{ l}\(\frac{\Gamma(l+d-1)}{\Gamma(l+1)\Gamma(d-1)}\)^{-\alpha/2}.
	\end{align}
	
	Let $a_l\defeq \(\frac{\Gamma(l+d-1)}{\Gamma(l+1)\Gamma(d-1)}\)^{-\alpha/2}\sqrt{l}.$ We first prove that when $d\ge \frac{10}{\alpha}2^{5/\alpha}+2$, $\frac{a_{l+1}}{a_l}\le \(\frac{l}{l+1}\)^2,\forall l\ge 1$. 
	By basic algebra we get
	\begin{align}
		\frac{a_{l+1}}{a_l}=\sqrt{\frac{l+1}{l}}\(\frac{l+1}{l+d-1}\)^{\alpha/2}.
	\end{align}
	Let $\kappa=2^{5/\alpha+1}$. We first focus on the case when $l\ge \frac{d}{\kappa-1}$. Since $\alpha(d-2)/5\ge \kappa$, we have
	\begin{align}
		\(\frac{l+1}{l+d-1}\)^{\alpha/5}=\(1-\frac{d-2}{l+d-1}\)^{\alpha/5}\le 1-\frac{\alpha(d-2)/5}{l+d-1}\le 1-\frac{\kappa}{l+d-1}.
	\end{align}
	When $l\ge \frac{d}{\kappa-1}$ we have $\frac{\kappa}{l+d-1}\ge \frac{1}{l+1}.$ As a result, $\(\frac{l+1}{l+d-1}\)^{\alpha/5}\le  1-\frac{1}{l+1}=\frac{l}{l+1}.$ Equivalently, we get
	\begin{align}
		\sqrt{\frac{l+1}{l}}\(\frac{l+1}{l+d-1}\)^{\alpha/2}\le \(\frac{l}{l+1}\)^{2}.\label{equ:truncation-2}
	\end{align}
	
	Now we focus on the case when $l<\frac{d}{\kappa-1}.$ In this case we have
	\begin{align}
		&\(\frac{l+1}{l+d-1}\)^{\alpha/2}< \(\frac{\frac{d}{\kappa-1}+1}{\frac{d}{\kappa-1}+d-1}\)^{\alpha/2}\le \(\frac{\frac{d}{\kappa-1}+2}{\frac{d}{\kappa-1}+d}\)^{\alpha/2}.
	\end{align}
	Since $\frac{d}{\kappa-1}\ge 2$, we have
	\begin{align}
		\(\frac{\frac{d}{\kappa-1}+2}{\frac{d}{\kappa-1}+d}\)^{\alpha/2}\le \(\frac{2\frac{d}{\kappa-1}}{\frac{d}{\kappa-1}+d}\)^{\alpha/2}=\(\frac{2}{\kappa}\)^{\alpha/2}\le 2^{5/2}\le \(\frac{l}{l+1}\)^{5/2}.
	\end{align}
	Consequently,
	\begin{align}
		\(\frac{l+1}{l+d-1}\)^{\alpha/2}\(\frac{l+1}{l}\)^{1/2}\le \(\frac{l}{l+1}\)^{2}.	\label{equ:truncation-3}
	\end{align}
	Combining Eq.~\eqref{equ:truncation-2} and Eq.~\eqref{equ:truncation-3}, in both cases we have
	\begin{align}
		\sqrt{\frac{l+1}{l}}\(\frac{l+1}{l+d-1}\)^{\alpha/2}\le \(\frac{l}{l+1}\)^{2}.
	\end{align}
	
	Now continue Eq.~\eqref{equ:truncation-4} we get,
	\begin{align}
		&\sum_{l\ge k}c\sqrt{l}\(\frac{\Gamma(l+d-1)}{\Gamma(l+1)\Gamma(d-1)}\)^{-\alpha/2}= ca_k\sum_{l\ge k}\frac{a_l}{a_k}=ca_k\sum_{l\ge k}\prod_{l'=k}^{l-1}\frac{a_{l'+1}}{a_{l'}}\\
		\le\;&ca_k\sum_{l\ge k}\frac{k^2}{l^2}\le cka_k=ck^{3/2}\(\frac{\Gamma(k+d-1)}{\Gamma(k+1)\Gamma(d-1)}\)^{-\alpha/2}\le ck^{3/2}2^{\alpha/2}(\Nkd)^{-\alpha/2}\\
		\le\;& 2ck^{3/2}(\Nkd)^{-\alpha/2}.
	\end{align}
\end{proof}

\subsection{Proof of Lemma~\ref{lem:main}}\label{app:pf-lem-main}
In this section we prove Lemma~\ref{lem:main}.

\begin{proof}[Proof of Lemma~\ref{lem:main}]
	Let $c=c_1c_2$. Recall that $k=\inf_{l\ge 0}\{2c(l+1)^{3/2}(N_{l+1,d})^{-\alpha/2}\le \epsilon/2\}.$ By Lemma~\ref{lem:truncation} we get
	\begin{align}\label{equ:thm-main-1}
		\|f-\Proj_{\le k}f\|_\infty\le \epsilon/2.
	\end{align}
	Hence, we only need to prove $\|\Proj_{\le k}g-\Proj_{\le k}f\|_\infty\le \epsilon/2$ and the desired result follows directly from triangle inequality.
	
	Since $\Proj_{\le k}(g-f)$ has degree at most $k$, applying Fact~\ref{fact:infty-2-SH} we get
	\begin{align}
		\|\Proj_{\le k}g-\Proj_{\le k}f\|_\infty\le \sum_{l=0}^{k}\|\Proj_{l}(g-f)\|_\infty\le \sqrt{\Nkd}\sum_{l=0}^{k}\|\Proj_{l}(g-f)\|_2.
	\end{align}
	By Cauchy-Schwarz and Parseval's theorem we have
	\begin{align}
		\sum_{l=0}^{k}\|\Proj_{l}(g-f)\|_2\le \((k+1)\sum_{l=0}^{k}\|\Proj_{l}(g-f)\|_2^2\)^{1/2}\le \sqrt{k+1}\|\Proj_{\le k}(g-f)\|_2.
	\end{align} As a result,
	\begin{align}\label{equ:thm-main-2}
		\|\Proj_{\le k}g-\Proj_{\le k}f\|_\infty\le \sqrt{k+1}\sqrt{\Nkd}\|\Proj_{\le k}(g-f)\|_2.
	\end{align}
	In the following, we show that
	\begin{align}
		\sqrt{k+1}\sqrt{\Nkd}\le 2(4c/\epsilon)^{3/\alpha}d^{4/\alpha}.
	\end{align}
	By the definition of $k$ we have $2ck^{3/2}(N_{k,d})^{-\alpha/2}> \epsilon/2.$
	Hence, 
	\begin{align}\label{equ:thm-main-3}
		\sqrt{N_{k,d}}\le \(\frac{4c}{\epsilon}k^{3/2}\)^{1/\alpha}.
	\end{align}
	To upper bound $k$, note that $N_{k,d}\ge (k/d)^{d-2}$. Therefore,
	\begin{align}
		\epsilon<4ck^{3/2}(N_{k,d})^{-\alpha/2}\le 4ck^{3/2} \(\frac{d}{k}\)^{-(d-2)\alpha/2}.
	\end{align}
	Solving for $k$ we get 
	$
	k\le \(4c/\epsilon\)^{\frac{2}{d\alpha-5}}d^{\frac{d\alpha-2}{d\alpha-5}}.
	$ Combining with Eq.~\eqref{equ:thm-main-3} and using the assumption that $d\ge \frac{10}{\alpha}2^{5/\alpha}+2$, we get
	\begin{align}
		&\sqrt{k+1}\sqrt{\Nkd}\le 2 \(4c/\epsilon\)^{\frac{1}{\alpha}}k^{\frac{3}{2\alpha}+\frac{1}{2}}\\
		\le\;& 2\(4c/\epsilon\)^{\frac{1}{\alpha}}(4c/\epsilon)^{\frac{2}{d\alpha-5}\(\frac{3}{2\alpha}+\frac{1}{2}\)}d^{\frac{d\alpha-2}{d\alpha-5}\(\frac{3}{2\alpha}+\frac{1}{2}\)}\le 2(4c/\epsilon)^{3/\alpha}d^{4/\alpha}.\label{equ:thm-main-4}
	\end{align}
	Finally, combining Eq.~\eqref{equ:thm-main-4}, Eq.~\eqref{equ:thm-main-2} and the assumption $\|\Proj_{\le k}(g-f)\|_2\le \frac{1}{4}\epsilon(4c/\epsilon)^{-3/\alpha}d^{-4/\alpha}$ we get
	\begin{align}
		\|\Proj_{\le k}g-\Proj_{\le k}f\|_\infty\le \epsilon/2.
	\end{align}
	By triangle inequality and Eq.~\eqref{equ:thm-main-1}, we prove the desired result:
	\begin{align}
		\|f-\Proj_{\le k}g\|_\infty\le \|f-\Proj_{\le k}f\|_\infty+ \|\Proj_{\le k}g-\Proj_{\le k}f\|_\infty\le \epsilon.
	\end{align}
\end{proof}

\subsection{Proof of Lemma~\ref{lem:relu-lb}}\label{app:pf-lem-relu-lb}
In this section we prove Lemma~\ref{lem:relu-lb}.
\begin{proof}[Proof of Lemma~\ref{lem:relu-lb}]
	In the following, we prove that for any $T<\Nkd$, there exists $f\in\calF_k$ such that $\Pr_{f,n}(\|\hat{f}_T-f\|_\infty<\beta_k/4)< 1/2$, and the desired result follows directly.
	
	Suppose at round $i$ the algorithm query $x_i\in \Sp$ and receive $y_i=f(x_i)+\calN(0,1)$ where $f$ is the ground-truth. At round $T$, the algorithm outputs $\hat{f}_T$. Let $\Pr_{u,n}(\cdot)$ be the probability space of $(x_1,y_1,\cdots,x_T,y_T)$ when the ground-truth is $f=\beta_k\Pkd(\dotp{\cdot}{u}),$ and $\Pr_{0,n}(\cdot)$ the space when the ground-truth is $f=0$. We use $\E_{u,n}$ and $\E_{0,n}$ to denote the corresponding expectation, respectively. Let $\calH_i$ be the $\sigma$-field of random variable $(x_1,y_1,\cdots,x_{i-1},y_{i-1},x_i).$
	
	For every $u\in\Sp$, let $E_{u,n}\defeq \ind{\|\hat{f}_T-\beta_k\Pkd(\dotp{\cdot}{u})\|_\infty<\beta_k/4}$ be the event that $\hat{f}_T$ is close to $\beta_k\Pkd(\dotp{\cdot}{u})$. By Pinsker's inequality and chain rule of KL divergence, we have
	\begin{align}
		\E_{u,n}[E_{u,n}]&\le \E_{0,n}[E_{u,n}]+\TV(\Pr_{0,n}\|\Pr_{u,n})\\
		&\le \E_{0,n}[E_{u,n}]+\sqrt{\frac{1}{2}\KL(\Pr_{0,n}\|\Pr_{u,n})}\\
		&= \E_{0,n}[E_{u,n}]+\sqrt{\frac{1}{2}\E_{0,n}\[\sum_{i=1}^{n}\KL(\Pr_{0,n}(y_i\mid \calH_i)\| \Pr_{u,n}(y_i\mid \calH_i))\]}\\
		&= \E_{0,n}[E_{u,n}]+\sqrt{\frac{\beta_k^2}{4}\E_{0,n}\[\sum_{i=1}^{n}\Pkd(x_i^\top u)^2\]}.
	\end{align}
	Consequently,
	\begin{align}
		\E_{u\sim \Sp}[\E_{u,n}[E_{u,n}]]&\le \E_{u\sim \Sp}\[\E_{0,n}[E_{u,n}]+\sqrt{\frac{\beta_k^2}{4}\E_{0,n}\[\sum_{i=1}^{n}\Pkd(x_i^\top u)^2\]}\]\\
		&\le \E_{u\sim \Sp}\[\E_{0,n}[E_{u,n}]\]+\sqrt{\frac{\beta_k^2}{4}\E_{u\sim \Sp}\[\E_{0,n}\[\sum_{i=1}^{n}\Pkd(x_i^\top u)^2\]\]}\\
		&= \E_{u\sim \Sp}\[\E_{0,n}[E_{u,n}]\]+\sqrt{\frac{\beta_k^2}{4}\E_{0,n}\[\sum_{i=1}^{n}\E_{u\sim \Sp}\[\Pkd(x_i^\top u)^2\]\]}\\
		&= \E_{u\sim \Sp}\[\E_{0,n}[E_{u,n}]\]+\sqrt{\frac{\beta_k^2}{4}\E_{0,n}\[\frac{n}{\Nkd}\]}\\
		&= \E_{u\sim \Sp}\[\E_{0,n}[E_{u,n}]\]+\sqrt{\frac{\beta_k^2}{4}\frac{n}{\Nkd}}.\label{equ:relu-lb-1}
	\end{align}
	
	Now we upper bound the first term in Eq.~\eqref{equ:relu-lb-1}. Let $\hat{u}_T=\min_{u\in\Sp}\|\hat{f}_T-\Pkd(\dotp{\cdot}{u})\|_\infty$. Consider the event $E_{u,n}'=\ind{\|\Pkd(\dotp{\cdot}{u})-\Pkd(\dotp{\cdot}{\hat{u}_n})\|_\infty <\beta_k/2}$. In the following we prove that $\neg E_{u,n}'\implies \neg E_{u,n}$. Indeed, when $\|\Pkd(\dotp{\cdot}{u})-\Pkd(\dotp{\cdot}{\hat{u}_n})\|_\infty \ge \beta_k/2$ we get
	\begin{align}
		\|\hat{f}_T-\Pkd(\dotp{\cdot}{u})\|_\infty&\ge \frac{1}{2}\(\|\hat{f}_T-\Pkd(\dotp{\cdot}{u})\|_\infty+\|\hat{f}_T-\Pkd(\dotp{\cdot}{\hat{u}_n})\|_\infty\)\tag{By the optimality of $\hat{u}_T$}\\
		&\ge\frac{1}{2}\|\Pkd(\dotp{\cdot}{u})-\Pkd(\dotp{\cdot}{\hat{u}_n})\|_\infty\ge \beta_k/4.\tag{Triangle inequality}
	\end{align}
	Therefore, we get
	\begin{align}
		&\E_{u\sim \Sp}\[\E_{0,n}[E_{u,n}]\]\le \E_{u\sim \Sp}\[\E_{0,n}[E_{u,n}']\]\\
		=\;&\E_{0,n}\[\E_{u\sim \Sp}\[\ind{\|\Pkd(\dotp{\cdot}{u})-\Pkd(\dotp{\cdot}{\hat{u}_n})\|_\infty\le \beta_k/4}\]\]\\
		\le \;&\E_{0,n}\[\E_{u\sim \Sp}\[\ind{|\Pkd(\dotp{u}{u})-\Pkd(\dotp{u}{\hat{u}_n})|\le \beta_k/4}\]\]\\
		= \;&\E_{0,n}\[\E_{u\sim \Sp}\[\ind{|1-\Pkd(\dotp{u}{\hat{u}_n})|\le \beta_k/4}\]\]\\
		\le \;&\E_{0,n}\[\Pr_{u\sim \Sp}\({\Pkd(\dotp{u}{\hat{u}_n})\ge 1-\beta_k/4}\)\]\\
		\le \;&\frac{16}{9\Nkd}\le \frac{1}{4}\tag{Proposition~\ref{prop:pnd-tail}}.
	\end{align}
	Finally, when $T<\Nkd \beta_k^{-2}$ we have
	\begin{align}
		\min_{f\in\calF_k}\Pr_{f,n}(\|\hat{f}_T-f\|_\infty<\beta_k/4)=\min_{u\in\Sp} \E_{u,n}[E_{u,n}]\le \E_{u\in\Sp}\E_{u,n}[E_{u,n}]<\frac{1}{2}.
	\end{align}
\end{proof}

\subsection{Proof of Theorem~\ref{thm:relu-lb}}\label{app:pf-thm-relu-lb}
In this section we present the proof of Theorem~\ref{thm:relu-lb}.
\begin{proof}[Proof of Theorem~\ref{thm:relu-lb}]
	When $\epsilon>d^{-7/4}$ the lower bound is trivial. Hence we focus on the regime $\epsilon<d^{-7/4}$.
	
	Let $k$ be the largest even number smaller than $\frac{d^{1/8}}{\sqrt{480\epsilon}}$ and $\tau_k=\dotp{\relu}{\Pkdb}_\mud$. First we prove that the set $\calF_k\defeq \{\tau_k\Pkd(\dotp{\cdot}{u}):u\in\Sp\}$ belongs to $\nnrelu(L_2).$
	
	To this end, we prove that for every $f\in\calF_k$, we can construct $c:\Sp\to\R$ such that $\|c\|_2\le 1$ and $f(x)=\E_{\xi\in\Sp}[\relu(\xi^\top x)c(\xi)]$ for every $x\in\Sp.$ For every $f=\tau_k\Pkd(\dotp{\cdot}{u})\in\calF_k$, by Funk-Hecke formula (Theorem~\ref{thm:funk-hecke}) we have
	\begin{align}
		f(x)=\tau_k\Pkd(\dotp{\cdot}{u})=\sqrt{\Nkd}\E_{\xi\in\Sp}[\relu(\xi^\top x)\Pkd(\dotp{\cdot}{u})]=\E_{\xi\in\Sp}[\relu(\xi^\top x)\Pkdb(\dotp{\cdot}{u})].
	\end{align}
	Since $\|\Pkdb(\dotp{\cdot}{u})\|_2=1$, we get $f\in \nnrelu(L_2).$
	
	In the following we prove the desired result by invoking Lemma~\ref{lem:relu-lb} with the hypothesis $\calF_k.$ First of all, by Lemma~\ref{lem:relu-eigen-approx} and the definition of $k$ we get
	\begin{align}
		\tau_k/4>\frac{d^{1/4}}{480k^{5/4}(k+d)^{3/4}}\ge \frac{d^{1/4}}{480k^2}\ge \epsilon.
	\end{align}
	Therefore, Lemma~\ref{lem:relu-lb} implies that the minimax sample complexity is at least $\Nkd\tau_k^{-2}.$ By basic Lemma~\ref{lem:relu-eigen-approx} and algebra we have
	\begin{align}
		&\Nkd\tau_k^{-2}\ge \binom{k+d-2}{d-2}\frac{k^{5/2}(k+d)^{3/2}}{1200 d^{1/2}}\gtrsim \(\frac{k}{d-2}+1\)^{d-2}\frac{k^{5/2}(k+d)^{3/2}}{ d^{1/2}}\\
		\ge&\(\frac{k}{d}\)^{d}=(0.002\epsilon^{-1}d^{-7/4})^{d/2},
	\end{align}
	which proves the desired result.
\end{proof}

\subsection{Missing Propositions}
In this section we state and prove the missing propositions in Section~\ref{sec:main-results}.

\begin{proposition}\label{prop:smooth-inner-product-decay}
	Let $h:[-1,1]\to \R$ be a one-dimensional function satisfies $\sup_{t\in[-1,1]}|h^{(k)}(t)|\le 1,\forall k\ge 0.$ Then 
	\begin{align}
		\|\Proj_k h(\dotp{\cdot}{u})\|_2\le 2\Nkd^{-1/2}.
	\end{align}
\end{proposition}
\begin{proof}
	For a fixed $u\in \Sp$, by the completeness of the Legendre polynomial basis, we have
	\begin{align}
		h(\dotp{\cdot}{u})=\sum_{k\ge 0} \tau_k \Pkdb(\dotp{\cdot}{u}),
	\end{align}
	where $\tau_k\defeq \dotp{h}{\Pkdb}_{\mud}$. Since $\Pkdb(\dotp{\cdot}{u})\in\Ykd$, it follows that 
	\begin{align}
		\|\Proj_k \exp(\dotp{\cdot}{u})\|_2=\tau_k \|\Pkdb(\dotp{\cdot}{u})\|_2=\tau_k.
	\end{align}
	As a result, we only need to prove that
	\begin{align}\label{equ:prop-sipd-1}
		\tau_k\le \Nkd^{-1/2},\quad \forall k\ge 0.
	\end{align}
	
	By Rodrigues formula \citet[Proposition 2.26]{atkinson2012spherical} we get
	\begin{align}
		\tau_k=\;&\int_{-1}^{1}h(t)\Pkdb(t)\mud(t)\dd t=\frac{\sqrt{\Nkd}\Gamma\(\frac{d}{2}\)}{\sqrt{\pi}\Gamma\(\frac{d-1}{2}\)}\int_{-1}^{1}h(t)\Pkd(t)(1-t^2)^{\frac{d-3}{2}}\dd t\\
		=\;&\frac{\sqrt{\Nkd}\Gamma\(\frac{d}{2}\)}{\sqrt{\pi}\Gamma\(\frac{d-1}{2}\)}\frac{\Gamma\(\frac{d-1}{2}\)}{2^k\Gamma\(k+\frac{d-1}{2}\)}\int_{-1}^{1}h^{(k)}(t)(1-t^2)^{k+\frac{d-3}{2}}\dd t\\
		\le \;&\frac{\sqrt{\Nkd}\Gamma\(\frac{d}{2}\)}{\sqrt{\pi}2^k\Gamma\(k+\frac{d-1}{2}\)}\int_{-1}^{1}|h^{(k)}(t)|(1-t^2)^{k+\frac{d-3}{2}}\dd t\\
		\le \;&\frac{\sqrt{\Nkd}\Gamma\(\frac{d}{2}\)}{\sqrt{\pi}2^k\Gamma\(k+\frac{d-1}{2}\)}\int_{-1}^{1}(1-t^2)^{k+\frac{d-3}{2}}\dd t\\
		\le \;&\frac{\sqrt{\Nkd}\Gamma\(\frac{d}{2}\)}{2^k\Gamma\(k+\frac{d-1}{2}\)}\frac{\Gamma\(k+\frac{d-1}{2}\)}{\Gamma\(k+\frac{d}{2}\)}=\frac{\sqrt{\Nkd}\Gamma\(\frac{d}{2}\)}{2^k\Gamma\(k+\frac{d}{2}\)}.
	\end{align}
	As a result, we only need to prove $\frac{\Gamma\(\frac{d}{2}\)}{2^k\Gamma\(k+\frac{d}{2}\)}\le 2\Nkd^{-1}$ and then Eq.~\eqref{equ:prop-sipd-1} follows directly.
	
	By the recursive formula of $\Gamma$ function we get 
	\begin{align}
		\frac{2^k\Gamma\(k+\frac{d}{2}\)}{\Gamma\(\frac{d}{2}\)}=2^k\prod_{l=1}^{k}\(k+\frac{d}{2}-l\)=\prod_{l=1}^{k}\(2k+d-2l\).
	\end{align}
	By the definition of $\Nkd$ we have
	\begin{align}
		\Nkd=\frac{2k+d-2}{k+d-2}\binom{k+d-2}{k}\le \frac{2\prod_{l=1}^{k}(k+d-1-l)}{k!}.
	\end{align}
	Observe that for any $l\in[1,k]$, $k+d-1-l\le 2k+d-2l.$ Consequently,
	\begin{align}
		\Nkd\le \frac{2\prod_{l=1}^{k}(k+d-1-l)}{k!}\le 2\prod_{l=1}^{k}\(2k+d-2l\)=2\frac{2^k\Gamma\(k+\frac{d}{2}\)}{\Gamma\(\frac{d}{2}\)}.
	\end{align}
	Equivalently,
	\begin{align}
		\frac{\Gamma\(\frac{d}{2}\)}{2^k\Gamma\(k+\frac{d}{2}\)}\le 2\Nkd^{-1}.
	\end{align}
\end{proof}

\begin{proposition}\label{prop:nn-exp-decay}
	Let $\nnexp(L_p)$ be the family of two layer NNs with activation $\exp(\cdot)$ and $L_p$ norm bounds. Then any function $f\in\nnexp(L_1)$ satisfies $\|\Proj_k f\|_2\le 2e\Nkd^{-1/2}.$
\end{proposition}
\begin{proof}
	Recall that if two functions $f,g$ satisfies $\|\Proj_k f\|_2\le \Nkd^{-1/2}$ and $\|\Proj_k g\|_2\le \Nkd^{-1/2}$, their convex combinations $h=\theta f+(1-\theta)g$ also satisfies $\|\Proj_k g\|_2\le \Nkd^{-1/2}$. Since any function in $\nnexp(L_1)$ can be written as a convex combination of functions $\{\pm \exp(\dotp{\cdot}{u}):u\in\Sp\}$, we only need to prove that 
	$\|\Proj_k \exp(\dotp{\cdot}{u})\|_2\le \Nkd^{-1/2}$ for every $u\in \Sp$.
	
	Let $h(t)=e^{-1}\exp(t)$. Then we have $\sup_{t\in[-1,1]}|h^{(k)}(t)|\le 1.$ Invoking Proposition~\ref{prop:smooth-inner-product-decay} we get
	\begin{align}
		\|\Proj_k \exp(\dotp{\cdot}{u})\|_2=e\|\Proj_k h(\dotp{\cdot}{u})\|_2\le 2e \Nkd^{-1/2}.
	\end{align}
\end{proof}

\begin{proposition}\label{prop:rkhs-norm-of-f}
	Suppose the function $f$ satisfies $\|\Proj_k f\|_2= \Omega(1)\Nkd^{-\alpha/2},\forall k\ge 0$ for some constant $\alpha>0$. For any inner product kernel $K(x,x')$ on the sphere where $\sup_{x,x'\in\Sp}|K(x,x')|\le 1$, $f$ has a infinite RKHS norm induced by $K$ when $\alpha<1/2$.
\end{proposition}
\begin{proof}
	Since $K(x,x')$ is a bounded inner product kernel, we can write $K(x,x')=h(\dotp{x}{x'})$ for some one-dimensional function $h:[-1,1]\to[-1,1].$ Let $\lambda_k$ be the eigenvalues of kernel $K$. By the Funk-Hecke formula (Theorem~\ref{thm:funk-hecke}) we get
	\begin{align}
		\lambda_k=\Nkd^{-1/2}\dotp{h}{\Pkdb}_{\mud}\le \Nkd^{-1/2}\|h\|_\mud\|\Pkdb\|_\mud\le \Nkd^{-1/2}\|h\|_\infty\|\Pkdb\|_\mud\le \Nkd^{-1/2}.
	\end{align}
	Since $\Ykd$ is the space of eigenfunctions of kernel $K$ corresponding to the eigenvalue $\lambda_k$, the RKHS norm of $f$ is defined by
	\begin{align}
		\|f\|_K^2=\sum_{k\ge 0}\frac{\|\Proj_k f\|_2^2}{\lambda_k}\ge \sum_{k\ge 0}\|\Proj_k f\|_2^2\Nkd^{1/2}.
	\end{align}
	As a result, when $\alpha<1/2$ we get
	\begin{align}
		\|f\|_K^2\gtrsim \sum_{k\ge 0}\Nkd^{1/2-\alpha}=\infty.
	\end{align}
\end{proof}
	\section{Learning with Two-layer Finite-width Neural Networks}\label{app:nns}
In this section, we show that using a finite-width two-layer neural network with polynomial activation can also achieve a small $L_\infty$-error bound.
\begin{algorithm}[ht]
	\caption{$L_\infty$-learning via Two-layer NNs with Polynomial Activation}
	\label{alg:NN}
	\hspace*{\algorithmicindent} \textbf{Input:} parameters $\alpha,c_1,c_2>0$, desired error level $\epsilon>0$, and failure probability $\delta>0.$
	
	\hspace*{\algorithmicindent} \textbf{Input:} Dataset $\calD=\{(x_i,y_i)\}_{i=1}^{n}$ where $x_i\sim \Sp$ are independent and uniformly sampled from the unit sphere $\Sp$, and $y_i=f(x_i)+\calN(0,1)$.
	\begin{algorithmic}[1]
		\State Set the truncation threshold $k\gets\inf_{l\ge 0}\{2c_1c_2(l+1)^{3/2}(N_{l+1,d})^{-\alpha/2}\le \epsilon/2\}.$\label{nn-line:3}
		\State Set the parameters for the neural network: norm bound $B=35 c_1\sqrt{d}\(\frac{4c_1c_2}{\epsilon}\)^{3+4/\alpha}$, and width $m\gets 256B^2\epsilon^{-6/\alpha-2}(4c_1c_2)^{6/\alpha}d^{8/\alpha}$.
		\State Define the family of two-layer NNs with polynomial activation $\sigma_k$ defined in Eq.~\eqref{equ:activation}:
		$$\textstyle{\calF_k=\left\{g(x)=\sum_{j=1}^{m}a_j\sigma_k(w_j^\top x): w_j\in \Sp,\sum_{j=1}^{m}|a_j|\le B\right\}}.$$
		\State Run empirical risk minimization and get
		$
		g=\textstyle{\argmin_{f\in\calF_k} \sum_{i=1}^{n}(f(x_i)-y_i)^2.}
		$
		\State \textbf{Return} $g$.
	\end{algorithmic}
\end{algorithm}

Our algorithm is presented as Alg.~\ref{alg:NN}. On a high level, given any desired error level $\epsilon>0$, the algorithm selects a truncation threshold $k\ge 0$ (Line~\ref{nn-line:3}), and use empirical risk minimization to find two-layer neural network $g$ with polynomial activation that fits the ground-truth $f$ the best. The activation $\sigma_k:[-1,1]\to\R$ is the degree-$k$ approximation of the ReLU activation in the Legendre polynomial space, given by
\begin{align}\label{equ:activation}
	\textstyle{\sigma_k(t)\defeq \sum_{l=0}^{k}\dotp{\relu}{\bar{P}_{l,d}}_\mud \bar{P}_{l,d}(t).}
\end{align}
Since $\Pkdb(\dotp{w}{\cdot})\in \Ykd$ for every $k\ge 0,w\in\Sp$, any two-layer NN with activation $\sigma_k$ is a degree-$k$ polynomial (more precisely, it is the projection of a two-layer ReLU network with the same parameters to the space $\mathbb{Y}_{\le k,d}$). Hence, our algorithm essentially aims to find the best low-degree approximation of the ground-truth $f$ using noisy data.

The following theorem states the sample complexity of Alg.~\ref{alg:NN}
\begin{theorem}\label{thm:nn-sc}
	Suppose the ground-truth function satisfies Conditions~\ref{cond:decay} and \ref{cond:random} for some fixed $\alpha\in (0,1]$ and $c_1,c_2>0$. If $d\ge 10\alpha^{-1}2^{5/\alpha}+2$, then for any $\epsilon>0,\delta>0$, with probability at least $1-\delta$ over the randomness of the data, Alg.~\ref{alg:NN} outputs a function $g$ such that $\|f-g\|_\infty\le \epsilon$ using $O(\poly(c_1c_2,d,1/\epsilon,\ln 1/\delta)^{1/\alpha})$ samples.
\end{theorem}

\subsection{Proof of Theorem~\ref{thm:nn-sc}}
\begin{proof}[Proof of Theorem~\ref{thm:nn-sc}]
	Let $\epsilon_1=\frac{1}{4}\epsilon^{3/\alpha+1}(4c_1c_2)^{-3/\alpha}d^{-4/\alpha}$. We prove Theorem~\ref{thm:mainsc} in the following two steps.
	\paragraph{Step 1: upper bound the population $L_2$ loss.} In this step, we use classic statistical learning tools to show that the ERM step (i.e., $g=\argmin_{h\in\calF_k} \sum_{i=1}^{n}(h(x_i)-y_i)^2$) returns a function $g$ with small $L_2$ loss. In particular, by Lemma~\ref{lem:ERM-l2-NN} we get
	$
	\|\Proj_{\le k} (f-g)\|_2\le \epsilon_1.
	$
	
	\paragraph{Step 2: upper bound the $L_\infty$-error via truncation.} This step is exactly the same as in the proof of Theorem~\ref{thm:mainsc}.
	
	Combining these two steps we prove the desired result.
\end{proof}

\begin{lemma}\label{lem:ERM-l2-NN}
	Suppose the function $f:\Sp\to \R$ satisfies Conditions~\ref{cond:decay} and \ref{cond:random} for some fixed $\alpha\in (0,1],c_1,c_2>0$. For any $\epsilon>0$, let $k=\inf_{l\ge 0}\{2c_1c_2(l+1)^{3/2}(N_{l+1,d})^{-\alpha/2}\le \epsilon/2\}$.
	
	For any $\epsilon_1>0$, let $B=35 c_1\sqrt{d}\(\frac{4c_1c_2}{\epsilon}\)^{3+4/\alpha}$, $\sigma_k(t)=\sum_{l=0}^{k}\dotp{\relu}{\bar{P}_{l,d}}_\mud \bar{P}_{l,d}(t)$, and $m=16B^2/\epsilon_1^2$, define the function class
	\begin{align}
		\calF_k=\left\{h(x)=\sum_{j=1}^{m}a_j\sigma_k(w_j^\top x): w_j\in \Sp,\sum_{j=1}^{m}|a_j|\le B\right\}.
	\end{align}
	For a given dataset $\{(x_i,y_i)\}_{i=1}^{n}$, let $\hat\calL(h)\defeq \frac{1}{N}\sum_{i=1}^{n}(h(x_i)-y_i)^2$ be the empirical $L_2$ loss, and $g=\argmin_{h\in\calF}\hat\calL(h).$
	
	For any $\delta>0$, when $d\ge \max\{2e,4/\alpha\}$ and $n\ge \Omega(\poly(d,(c_1c_2)^{1/\alpha},\epsilon^{-1/\alpha},\ln(1/\delta),1/\epsilon_1))$, with probability at least $1-\delta$,
	\begin{align}
		\|\Proj_{\le k} (f-g)\|_2=\|\Proj_{\le k} f-g\|_2\le \epsilon_1.
	\end{align}
\end{lemma}

\subsection{Proof of Lemma~\ref{lem:ERM-l2-NN}}\label{app:pf-ERM-l2-NN}
In this section, we prove Lemma~\ref{lem:ERM-l2-NN}.
\begin{proof}[Proof of Lemma~\ref{lem:ERM-l2-NN}]
	First we prove that there exists $\hat{f}\in\calF$ such that the population loss is small. Since $\calF\subseteq \mathbb{Y}_{\le k,d}$, we get
	\begin{align}\label{equ:ERM-l2-0}
		\forall h\in\calF, \quad \|h-f\|_2^2=\|h-\Proj_{\le k}f\|_2^2+\|f-\Proj_{\le k}f\|_2^2.
	\end{align}
	
	By Lemma~\ref{lem:truncated-f-l1-norm}, $\Proj_{\le k}f$ can be represented by a infinite-width two-layer ReLU neural network with weight $c$ such that $\|c\|_1\le 35 c_1\sqrt{d}\(\frac{4c_1c_2}{\epsilon}\)^{3+4/\alpha}=B$. By Lemma~\ref{lem:nn-sample}, when $m>16B^2/\epsilon_1^2$ there exists a finite-width approximation $\hat{f}\in\calF$ such that $\|\hat{f}-\Proj_{\le k}f\|_2\le \epsilon_1/2.$ 
	
	In the following we show that ERM outputs a function $g\in\calF$ such that 
	\begin{align}\label{equ:ERM-l2-1}
		\|g-f\|_2^2\le \|\hat{f}-f\|_2^2+\epsilon_1^2/2.
	\end{align}
	By the uniform convergence of two-layer neural networks (Lemma~\ref{lem:uniform-convergence}), when $$n\ge \Omega(\poly(d,(c_1c_2)^{1/\alpha},\epsilon^{-1/\alpha},\ln(1/\delta),1/\epsilon_1))$$ we have
	\begin{align}
		&\|g-f\|_2^2\le \hat\calL(g)+\epsilon_1^2/4\le \hat\calL(\hat{f})+\epsilon_1^2/4\le\|\hat{f}-f\|_2^2+\epsilon_1^2/2.
	\end{align}
	Combining with Eq.~\eqref{equ:ERM-l2-0} we get
	\begin{align}
		\|g-\Proj_{\le k} f\|_2^2\le \|\hat{f}-\Proj_{\le k} f\|_2^2+\epsilon_1^2/2< \epsilon_1^2.
	\end{align}
\end{proof}

The following lemma proves the uniform convergence result for the function class used in Lemma~\ref{lem:ERM-l2-NN}.
\begin{lemma}\label{lem:uniform-convergence}
	In the setting of Lemma~\ref{lem:ERM-l2-NN}, when $n\ge \Omega(\poly(B,\Nkd,\ln(1/\delta),1/\epsilon_1))$, for any $\delta>0$, with probability at least $1-\delta$ we have
	\begin{align}
		\sup_{g\in\calF}|\|g-f\|_2^2-\hat\calL(g)|\le \epsilon_1.
	\end{align}
\end{lemma}
\begin{proof}
	The proof is essentially the same as the proof of Lemma~\ref{lem:uniform-convergence-kernel}, with the only difference that here we use the Rademacher complexity upper bound for two-layer neural networks~\citep[Theorem 18]{bartlett2002rademacher}.
\end{proof}

The following lemma proves the realizability result for the function class used in Lemma~\ref{lem:ERM-l2-NN}.
\begin{lemma}\label{lem:truncated-f-l1-norm}
	In the setting of Lemma~\ref{lem:ERM-l2-NN}, $\Proj_{\le k}f$ can be represented by an infinite-width two-layer ReLU neural network with weight $c:\Sp\to\R$ such that $\|c\|_1\le 35 c_1\sqrt{d}\(\frac{4c_1c_2}{\epsilon}\)^{3+4/\alpha}$.
\end{lemma}
\begin{proof}
	Recall that we can write $\Proj_{\le k} f(x)=\sum_{l=0}^{k} \sum_{j=1}^{\Nkd}a_{l,j}Y_{l,j}(\cdot)$. 
	Let \begin{align}
		\lambda_l=\Nld^{-1/2}\dotp{\relu}{\Pldb}_\mud
	\end{align} and 
	define the weight $c:\Sp\to\R$ by $c(x)=\sum_{l=0}^{k}\lambda_l^{-1}\sum_{j=1}^{\Nld}a_{l,j}Y_{l,j}(\cdot).$ Then by the Funk-Hecke formula (Theorem~\ref{thm:funk-hecke}) we get
	\begin{align}
		\Proj_{\le k} f(x)=\E_{w\sim\Sp}[\sigma(x^\top w)c(w)],\quad\forall x\in\Sp.
	\end{align}
	Hence, we only need to upper bound $\|c\|_2,$ and then the desired result is proved by the fact that $\|c\|_1\le \|c\|_2.$
	
	Let $\beta_l^2=\|\Proj_l f\|_2^2=\sum_{j=1}^{\Nld}a_{l,j}^2$ for all $l\in[0,k]$. Then we have
	\begin{align}
		&\|c\|_2^2=\sum_{l=0}^{k}\lambda_l^{-2}\beta_l^2\le \sum_{l=0}^{k}\lambda_l^{-2}c_1\Nld^{-\alpha}
		\le1200c_1^2\sum_{l=0}^{k}\Nld^{1-\alpha}l(l+d) \tag{By Lemma~\ref{lem:relu-eigen-approx}}\\
		\le\;&1200c_1^2\Nkd k^2(k+d).
	\end{align}
	By the definition of $k$ we have
	\begin{align}
		2c_1c_2k^{3/2}(\Nkd)^{-\alpha/2}> \epsilon/2.
	\end{align}
	Consequently,
	\begin{align}
		\Nkd< \(\frac{4c_1c_2}{\epsilon}\)^{2/\alpha}k^{3/\alpha}.
	\end{align}
	Applying Proposition~\ref{prop:truncation-upperbound} we get $k\le \(\frac{4c_1c_2}{\epsilon}\)^{\frac{2}{d\alpha-3}}$. Using the assumption that $d>4/\alpha$ we get
	\begin{align}
		c_1^2\Nkd k^2(k+d)\le dc_1^2\(\frac{4c_1c_2}{\epsilon}\)^{2/\alpha}k^{3+3/\alpha}
	\end{align}
	As a result,
	\begin{align}
		\|c\|_1^2\le \|c\|_2^2\le 1200 dc_1^2\(\frac{4c_1c_2}{\epsilon}\)^{6+8/\alpha}.
	\end{align}
\end{proof}
	\section{Decomposition of ReLU in the Legendre Polynomial Space}
The following lemma analytically computes the spherical harmonics decomposition of ReLU activation (see also \citet{bach2017breaking,mhaskar2006weighted,bourgain1988projection,schneider1967problem}).
\begin{lemma}\label{lem:relu-eigen}
	Let $\tau_k=\dotp{\relu}{\bar{P}_{k,d}}_\mud$ be the projection of ReLU function to degree-$k$ Legendre polynomial. Then we have
	\begin{align}
		\tau_k=\begin{cases}
			(-1)^{\frac{k-2}{2}}\sqrt{\Nkd}\frac{1}{2^k\sqrt{\pi}}\frac{\Gamma(d/2)\Gamma(k-1)}{\Gamma(k/2)\Gamma((k+d+1)/2)},&\text{when $k$ is even},\\
			\frac{1}{2\sqrt{d}},&\text{when $k=1$},\\
			0,&\text{when $k>1$ and $k$ is odd}.
		\end{cases}
	\end{align}
\end{lemma}
\begin{proof}
	Recall that by definition,
	\begin{align}\label{equ:tau-def}
		\tau_k=\int_{-1}^{1}\relu(t)\bar{P}_{k,d}(t)\mu_d(t)\dd t=\sqrt{\Nkd}\int_{0}^{1}t P_{k,d}(t)\mu_d(t)\dd t.
	\end{align}
	When $k$ is odd we have $P_{k,d}(-t)=-P_{k,d}(t).$ As a result,
	\begin{align}
		\int_{0}^{1}t P_{k,d}(t)\mu_d(t)\dd t=\frac{1}{2}\int_{-1}^{1}t P_{k,d}(t)\mu_d(t)\dd t.
	\end{align}
	Recall that $P_{1,d}(t)=t$, and we have $$\int_{-1}^{1}t P_{k,d}(t)\mu_d(t)\dd t=\int_{-1}^{1}P_{1,d}(t) P_{k,d}(t)\mu_d(t)\dd t=\frac{1}{\Nkd}\ind{k=1}.$$ It follows directly that (1) $\tau_k=0$ if $k>1$ and $k$ is odd, and (2) $\tau_1=\frac{1}{2\sqrt{N_{1,d}}}=\frac{1}{2\sqrt{d}}.$
	
	Now we focus on the case when $k$ is even. By the Rodrigues representation formula \citep[Theorem 2.23]{atkinson2012spherical} we get
	\begin{align}
		P_{k,d}(t)=(-1)^k\frac{\Gamma(\frac{d-1}{2})}{2^k\Gamma(k+\frac{d-1}{2})}(1-t^2)^{-\frac{d-3}{2}}\(\frac{\dd}{\dd t}\)^k(1-t^2)^{k+\frac{d-3}{2}}.
	\end{align}
	As a result,
	\begin{align}
		&\int_{0}^{1}t P_{k,d}(t)\mu_d(t)\dd t\\
		=&(-1)^k\frac{\Gamma(\frac{d-1}{2})}{2^k\Gamma(k+\frac{d-1}{2})}\frac{\Gamma(d/2)}{\Gamma((d-1)/2)}\frac{1}{\sqrt{\pi}}\int_{0}^{1}t\(\frac{\dd}{\dd t}\)^k(1-t^2)^{k+\frac{d-3}{2}}\dd t\\
		=&(-1)^{k+1}\frac{\Gamma(\frac{d-1}{2})}{2^k\Gamma(k+\frac{d-1}{2})}\frac{\Gamma(d/2)}{\Gamma((d-1)/2)}\frac{1}{\sqrt{\pi}}\int_{0}^{1}\(\frac{\dd}{\dd t}\)^{k-1}(1-t^2)^{k+\frac{d-3}{2}}\dd t\tag{integration by parts}\\
		=&(-1)^{k+1}\frac{\Gamma(\frac{d-1}{2})}{2^k\Gamma(k+\frac{d-1}{2})}\frac{\Gamma(d/2)}{\Gamma((d-1)/2)}\frac{1}{\sqrt{\pi}}\(\frac{\dd}{\dd t}\)^{k-2}(1-t^2)^{k+\frac{d-3}{2}}\bigg\vert_{0}^{1}\\
		=&(-1)^{k}\frac{\Gamma(\frac{d-1}{2})}{2^k\Gamma(k+\frac{d-1}{2})}\frac{\Gamma(d/2)}{\Gamma((d-1)/2)}\frac{1}{\sqrt{\pi}}\(\frac{\dd}{\dd t}\)^{k-2}(1-t^2)^{k+\frac{d-3}{2}}\bigg\vert_{t=0}.\label{equ:relu-pf-1}
	\end{align}
	By binomial theorem, we have
	\begin{align}
		&\(\frac{\dd}{\dd t}\)^{k-2}(1-t^2)^{k+\frac{d-3}{2}}\bigg\vert_{t=0}=\(\frac{\dd}{\dd t}\)^{k-2} \sum_{j=0}^{k+\frac{d-3}{2}}{{k+\frac{d-3}{2}}\choose j}(-1)^jt^{2j}\bigg\vert_{t=0}\\
		=&(-1)^{\frac{k-2}{2}}(k-2)!{{k+\frac{d-3}{2}}\choose \frac{k-2}{2}}=(-1)^{\frac{k-2}{2}}\frac{\Gamma(k-1)\Gamma(k+\frac{d-1}{2})}{\Gamma(k/2)\Gamma(\frac{k+d+1}{2})}.\label{equ:relu-pf-2}
	\end{align}
	Combining Eq.~\eqref{equ:relu-pf-1} and Eq.~\eqref{equ:relu-pf-2} we get
	\begin{align}
		&\int_{0}^{1}t P_{k,d}(t)\mu_d(t)\dd t\\
		=&(-1)^{k}\frac{\Gamma(\frac{d-1}{2})}{2^k\Gamma(k+\frac{d-1}{2})}\frac{\Gamma(d/2)}{\Gamma((d-1)/2)}\frac{1}{\sqrt{\pi}}(-1)^{\frac{k-2}{2}}\frac{\Gamma(k-1)\Gamma(k+\frac{d-1}{2})}{\Gamma(k/2)\Gamma(\frac{k+d+1}{2})}\\
		=&(-1)^{\frac{k-2}{2}}\frac{\Gamma(\frac{d}{2})\Gamma(k-1)}{2^k\Gamma(\frac{k}{2})\Gamma(\frac{k+d+1}{2}))}\frac{1}{\sqrt{\pi}}.
	\end{align}
	Finally, combining with Eq.~\eqref{equ:tau-def} we prove the desired result.
\end{proof}

\begin{lemma}\label{lem:relu-eigen-approx}
	Let $\tau_k=\dotp{\relu}{\bar{P}_{k,d}}_\mud$ be the projection of ReLU to degree-$k$ Legendre polynomial. Then we have $\abs{\tau_k}=\Theta(d^{1/4}k^{-5/4}(k+d)^{-3/4}).$ In particular, for all dimension $d\ge 3$ and even degree $k\ge 4$ the following upper and lower bounds hold:
	\begin{align}
		\frac{2^{5/4}\pi^{3/4}}{\exp(13/2)} d^{1/4}k^{-5/4}(k+d)^{-3/4}\le \abs{\tau_k}\le \frac{\exp(13/2)}{2\pi^2} d^{1/4}k^{-5/4}(k+d)^{-3/4}.
	\end{align}
\end{lemma}
\begin{proof}
	Recall that Stirling's formula states
	\begin{align}
		\sqrt{2\pi}k^{k+1/2}e^{-k}\le \Gamma(k+1)\le e k^{k+1/2}e^{-k}.
	\end{align}
	We first prove the upper bound. By Lemma~\ref{lem:relu-eigen}, when $k$ is even we have
	\begin{align}
		\abs{\tau_k}=&\;\sqrt{\Nkd}\frac{1}{2^k\sqrt{\pi}}\frac{\Gamma(d/2)\Gamma(k-1)}{\Gamma(k/2)\Gamma((k+d+1)/2)}\\
		=&\;\sqrt{\frac{2k+d-2}{k+d-2}\frac{\Gamma(k+d-1)}{\Gamma(k+1)\Gamma(d-1)}}\frac{1}{2^k\sqrt{\pi}}\frac{\Gamma(d/2)\Gamma(k-1)}{\Gamma(k/2)\Gamma((k+d+1)/2)}\\
		\le &\;\frac{\sqrt{2}}{\sqrt{\pi}}\(\sqrt{\frac{\Gamma(k+d-1)}{\Gamma(k+1)\Gamma(d-1)}}\frac{1}{2^k}\frac{\Gamma(d/2)\Gamma(k-1)}{\Gamma(k/2)\Gamma((k+d+1)/2)}\)\\
		\le &\;\frac{\sqrt{2}}{\sqrt{\pi}}\(\sqrt{\frac{e}{2\pi}\frac{(k+d-2)^{k+d-\frac{3}{2}}}{k^{k+\frac{1}{2}}(d-2)^{d-\frac{3}{2}}}}\frac{1}{2^k}\frac{\exp(7/2)}{2\pi}\frac{(\frac{d}{2}-1)^{\frac{d-1}{2}}(k-2)^{k-\frac{3}{2}}}{(\frac{k}{2}-1)^{\frac{k-1}{2}}(\frac{k+d-1}{2})^{\frac{k+d}{2}}}\)\\
		\le &\;\frac{\exp(4)}{2\pi^2}\(\exp(5/2)\sqrt{\frac{(k+d)^{k+d-\frac{3}{2}}}{k^{k+\frac{1}{2}}d^{d-\frac{3}{2}}}}\frac{1}{2^k}\frac{(\frac{d}{2})^{\frac{d-1}{2}}k^{k-\frac{3}{2}}}{(\frac{k}{2})^{\frac{k-1}{2}}(\frac{k+d}{2})^{\frac{k+d}{2}}}\)\tag{Since $(1-1/t)^t=\Theta(1)$}\\
		\le &\;\frac{\exp(13/2)}{2\pi^2}\(\sqrt{\frac{(k+d)^{k+d-\frac{3}{2}}}{k^{k+\frac{1}{2}}d^{d-\frac{3}{2}}}}\frac{d^{\frac{d-1}{2}}k^{k-\frac{3}{2}}}{k^{\frac{k-1}{2}}(k+d)^{\frac{k+d}{2}}}\)\\
		\le&\;\frac{\exp(13/2)}{2\pi^2}\((k+d)^{-3/4}k^{-5/4}d^{1/4}\).
	\end{align}
	Now we prove the lower bound. Similarly,
	\begin{align}
		\abs{\tau_k}=&\;\sqrt{\Nkd}\frac{1}{2^k\sqrt{\pi}}\frac{\Gamma(d/2)\Gamma(k-1)}{\Gamma(k/2)\Gamma((k+d+1)/2)}\\
		=&\;\sqrt{\frac{2k+d-2}{k+d-2}\frac{\Gamma(k+d-1)}{\Gamma(k+1)\Gamma(d-1)}}\frac{1}{2^k\sqrt{\pi}}\frac{\Gamma(d/2)\Gamma(k-1)}{\Gamma(k/2)\Gamma((k+d+1)/2)}\\
		\ge &\;\frac{1}{\sqrt{\pi}}\(\sqrt{\frac{\Gamma(k+d-1)}{\Gamma(k+1)\Gamma(d-1)}}\frac{1}{2^k}\frac{\Gamma(d/2)\Gamma(k-1)}{\Gamma(k/2)\Gamma((k+d+1)/2)}\)\\
		\ge &\;\frac{1}{\sqrt{\pi}}\(\sqrt{\frac{\sqrt{2\pi}}{e^2}\frac{(k+d-2)^{k+d-\frac{3}{2}}}{k^{k+\frac{1}{2}}(d-2)^{d-\frac{3}{2}}}}\frac{1}{2^k}\frac{2\pi}{\exp(1/2)}\frac{(\frac{d}{2}-1)^{\frac{d-1}{2}}(k-2)^{k-\frac{3}{2}}}{(\frac{k}{2}-1)^{\frac{k-1}{2}}(\frac{k+d-1}{2})^{\frac{k+d}{2}}}\)\\
		\ge &\;\frac{2^{5/4}\pi^{3/4}}{\exp(3/2)}\(\exp(-5)\sqrt{\frac{(k+d)^{k+d-\frac{3}{2}}}{k^{k+\frac{1}{2}}d^{d-\frac{3}{2}}}}\frac{1}{2^k}\frac{(\frac{d}{2})^{\frac{d-1}{2}}k^{k-\frac{3}{2}}}{(\frac{k}{2})^{\frac{k-1}{2}}(\frac{k+d}{2})^{\frac{k+d}{2}}}\)\tag{Since $(1-1/t)^t=\Theta(1)$}\\
		\ge &\;\frac{2^{5/4}\pi^{3/4}}{\exp(13/2)}\(\sqrt{\frac{(k+d)^{k+d-\frac{3}{2}}}{k^{k+\frac{1}{2}}d^{d-\frac{3}{2}}}}\frac{d^{\frac{d-1}{2}}k^{k-\frac{3}{2}}}{k^{\frac{k-1}{2}}(k+d)^{\frac{k+d}{2}}}\)\\
		\ge&\;\frac{2^{5/4}\pi^{3/4}}{\exp(13/2)}\((k+d)^{-3/4}k^{-5/4}d^{1/4}\).
	\end{align}
\end{proof}

\section{Random Spherical Harmonics}\label{app:random-sh}
In this section, we prove the $L_\infty$-norm bound for random spherical harmonics. \citet{burq2014probabilistic} prove a similar result without explicitly computes the $d$-dependency in Eq.~\eqref{equ:random-sh}.
\begin{proof}[Proof of Lemma~\ref{lem:random-sh}]
	For any fixed $x\in\Sp$, by Lemma~\ref{lem:riesz-SH} we get
	\begin{align}\label{equ:random-sh-1}
		g(x)=\sqrt{\Nkd}\E_{\xi\sim \Sp}[g(\xi)\Pkdb(x^\top \xi)].
	\end{align}
	Since $\{Y_{k,j}\}_{j=1}^{\Nkd}$ is a set of orthonormal basis, there exists weights $\{u_j\}_{j=1}^{\Nkd}$ (that depends on $x$) such that 
	\begin{align}
		\Pkdb(x^\top \xi)=\sum_{j=1}^{\Nkd}u_jY_{k,j}(\xi),\quad \forall \xi\in\Sp,
	\end{align}
	and $\sum_{j=1}^{\Nkd}u_j^2=\E_{\xi\sim \Sp}[\Pkdb(x^\top \xi)^2]=1.$
	Define $a=[a_1,\cdots,a_{\Nkd}]\in\R^{\Nkd}$ and $u=[u_1,\cdots,u_{\Nkd}]\in\R^{\Nkd}$. Then we have
	\begin{align}
		&g(x)=\sqrt{\Nkd}\E_{\xi\sim \Sp}[g(\xi)\Pkdb(x^\top \xi)]=\sqrt{\Nkd}\E_{\xi\sim \Sp}\[\(\sum_{j=1}^{\Nkd}a_{j}Y_{k,j}(\xi)\)\(\sum_{j=1}^{\Nkd}u_jY_{k,j}(\xi)\)\]\nonumber\\
		=\;&\sqrt{\Nkd}\sum_{j=1}^{\Nkd}a_{j}u_j=\sqrt{N_{k,d}}a^\top u.
	\end{align}
	In addition, $\|g\|_2^2=\sum_{j=1}^{\Nkd}a_j^2=\|a\|_2^2$. Hence, by Lemma~\ref{lem:gaussian-proj-concentration} we get 
	\begin{align}
		\forall t>0,\Pr\(\frac{|a^\top u|}{\|a\|_2}\ge \frac{2t}{\sqrt{\Nkd}}\)\le 3\exp(-t^2/2).
	\end{align} Equivalently,
	\begin{align}\label{equ:random-sh-3}
		\forall x\in\Sp,t>0,\quad \Pr\(|g(x)|\ge 2t\|g\|_2 \)\le 3\exp(-t^2/2).
	\end{align}
	In the following, we upper bound $|g(x)|/\|g\|_2$ uniformly over all $x\in\Sp$ by the covering number argument and uniform concentration.
	
	Let $h(x)=g(x)/\|g\|_2$. First we prove that $h(x)$ is Lipschitz on $\Sp$ with respect to the great-circle distance $d(x,y)\defeq \arccos(x^\top y).$ To this end, we only need to upper bound the manifold gradient $\nabla^\star_x h(x)$ on the sphere. By Eq.~\eqref{equ:random-sh-1} we get,
	\begin{align}
		&\|\nabla^\star_x h(x)\|_2=\frac{1}{\|g\|_2}\sqrt{\Nkd}\|\E_{\xi\sim \Sp}[g(\xi)\nabla^\star_x \Pkdb(x^\top \xi)]\|_2\\
		\le\;&\frac{1}{\|g\|_2}\sqrt{\Nkd}\E_{\xi\sim \Sp}[|g(\xi)| \|\nabla^\star_x \Pkdb(x^\top \xi)\|_2]\\
		\le\;&\frac{1}{\|g\|_2}\sqrt{\Nkd}\(\E_{\xi\sim \Sp}[g(\xi)^2] \E_{\xi\sim \Sp}[\|\nabla^\star_x \Pkdb(x^\top \xi)\|_2^2]\)^{1/2}\tag{Cauchy-Schwarz inequality}\\
		\le\;&\sqrt{\Nkd}\sqrt{k(k+d-2)}.\tag{\citet[Proposition 3.6]{atkinson2012spherical}}
	\end{align}
	which implies that $h(x)$ is $(\sqrt{k(k+d-2)\Nkd})$-Lipschitz.
	
	Let $\epsilon=(2\sqrt{k(k+d-2)\Nkd})^{-1}$ and $\calC$ an $\epsilon$-covering of $\Sp$ with respect to the great-circle distance. By Proposition~\ref{prop:covering-sphere} we get $|\calC|\le (3/\epsilon)^d$. In addition, for every $x\in\Sp$ there exists $\hat{x}\in \calC$ such that 
	\begin{align}\label{equ:random-sh-4}
		|h(x)-h(\hat{x})|\le \sqrt{k(k+d-2)\Nkd}\epsilon=\frac{1}{2}.
	\end{align}
	By union bound and Eq.~\eqref{equ:random-sh-3}, with probability at least $1-\delta$ we get,
	\begin{align}
		\forall x\in\calC, \quad |h(x)|&\le 4\sqrt{\ln \frac{3|\calC|}{\delta}}\le 4\sqrt{\ln (3/\delta)+d\ln (3/\epsilon)}\\
		&\le 4\sqrt{\ln(3/\delta)+2d^2\ln(k+1)}.
	\end{align}
	Combining with Eq.~\eqref{equ:random-sh-4} we get, with probability at least $1-\delta$,
	\begin{align}
		\forall x\in\Sp,\quad \frac{|g(x)|}{\|g\|_2}\le 5\sqrt{\ln(3/\delta)+2d^2\ln(k+1)},
	\end{align}
	which proves the desired result.
\end{proof}

The following lemma is an realization of Riesz representation theorem.
\begin{lemma}\label{lem:riesz-SH}
	For any fixed $k\ge 0$ and $f\in\Ykd$, we have
	\begin{align}
		f(x)=\sqrt{\Nkd}\E_{\xi \sim \Sp}[f(\xi)\Pkdb(x^\top \xi)],\quad \forall x\in \Sp.
	\end{align}
\end{lemma}
\begin{proof}
	let $\{Y_{k,j}\}_{j=1}^{\Nkd}$ be any set of orthonormal basis for degree $k$ spherical harmonics $\Ykd$. By addition theorem \citet[Theorem 2.9]{atkinson2012spherical}, for any $\xi\in\Sp$ we get
	\begin{align}
		\sqrt{\Nkd}\Pkdb(x^\top \xi)=\sum_{j=1}^{\Nkd}Y_{k,j}(x)Y_{k,j}(\xi).
	\end{align}
	Since $\{Y_{k,j}\}_{j=1}^{\Nkd}$ is a set of orthonormal basis, there exists unique coefficients $\{a_j\}_{j=1}^{\Nkd}$ such that $
		f(x)=\sum_{j=1}^{\Nkd}a_{j}Y_{k,j}(x),\quad \forall x\in\Sp.
	$ As a result,
	\begin{align}
		&\sqrt{\Nkd}\E_{\xi \sim \Sp}[f(\xi)\Pkdb(x^\top \xi)]=\E_{\xi \sim \Sp}\[\(\sum_{j=1}^{\Nkd}a_{j}Y_{k,j}(\xi)\)\(\sum_{j=1}^{\Nkd}Y_{k,j}(x)Y_{k,j}(\xi)\)\]\notag\\
		=\;&\sum_{j=1}^{\Nkd}a_{j}Y_{k,j}(x)=f(x).
	\end{align}
\end{proof}

\section{Helper Lemmas}
In this section, we present some low-level helper lemmas.

\begin{proposition}\label{prop:covering-sphere}
	Let $N(\epsilon)$ be the $\epsilon$-covering number of $\Sp$ with respect to the great-circle distance $d(x,y)\defeq \arccos(x^\top y).$ Then for any $\epsilon\in(0,1)$ we have
	\begin{align}
		N(\epsilon)\le (3/\epsilon)^d.
	\end{align}
\end{proposition}
\begin{proof}
	Note that any $(\epsilon/2)$-cover of the unit ball $B^{d}$ (w.r.t. the Euclidean distance) induces an $\epsilon$-covering of the unit sphere $\Sp$ with the same size. As a result,
	\begin{align}
		N(\epsilon)\le \frac{(1+\epsilon/2)^d}{(\epsilon/2)^d}\le \(\frac{3}{\epsilon}\)^d.
	\end{align}
\end{proof}

\begin{proposition}\label{prop:bessel}
	Let $I_\nu(z)\defeq\sum_{j=0}^{\infty}\frac{1}{j!\Gamma(\nu+j+1)}\(\frac{z}{2}\)^{\nu+2j}$ be the modified Bessel function of the first kind. Then for every $\nu>1$ we get
	\begin{align}
		\sqrt{2}e^{-1}\frac{e^{\nu}}{(2\nu)^{\nu+1/2}} \le I_{\nu}(1)\le \frac{e^{1/4}}{\sqrt{\pi}}\frac{e^{\nu}}{(2\nu)^{\nu+1/2}}.
	\end{align}
\end{proposition}
\begin{proof}
	By algebraic manipulation we get,
	\begin{align}
		I_\nu(1)=\sum_{j=0}^{\infty}\frac{1}{j!\Gamma(\nu+j+1)}\(\frac{1}{2}\)^{\nu+2j}
		\le \frac{(1/2)^\nu}{\Gamma(\nu+1)}\sum_{j=0}^{\infty}\frac{(1/2)^{2j}}{j!}= \frac{(1/2)^\nu}{\Gamma(\nu+1)}e^{1/4}.
	\end{align}
	By Stirling's formula, we get
	\begin{align}
		\frac{(1/2)^\nu}{\Gamma(\nu+1)}e^{1/4}\le \frac{(1/2)^\nu e^\nu}{\sqrt{2\pi}\nu^{\nu+1/2}}e^{1/4}=\frac{e^{1/4}}{\sqrt{\pi}}\frac{e^{\nu}}{(2\nu)^{\nu+1/2}}.
	\end{align}
	Similarly, we have
	\begin{align}
		I_\nu(1)=\sum_{j=0}^{\infty}\frac{1}{j!\Gamma(\nu+j+1)}\(\frac{1}{2}\)^{\nu+2j}\ge \frac{(1/2)^\nu}{\Gamma(\nu+1)}.
	\end{align}
	Using Stirling's formula again we have,
	\begin{align}
		\frac{(1/2)^\nu}{\Gamma(\nu+1)}\ge \frac{(1/2)^\nu e^\nu}{e\nu^{\nu+1/2}}=\sqrt{2}e^{-1}\frac{e^{\nu}}{(2\nu)^{\nu+1/2}}.
	\end{align}
\end{proof}

\begin{proposition}\label{prop:pnd-tail}
	For any fixed $k\ge 0$, $u\in\Sp$, and $t>0$ we have
	\begin{align}
		\Pr_{x\sim \Sp}(|\Pkd(x^\top u)|>t)\le \frac{1}{t^2\Nkd}.
	\end{align}
\end{proposition}
\begin{proof}
	By Markov ineqaulity we have
	\begin{align}
		&\Pr_{x\sim \Sp}(|\Pkd(x^\top u)|>t)=\Pr_{x\sim \Sp}(\Pkd(x^\top u)^2>t^2)\\
		\le\; &t^{-2}\E_{x\sim \Sp}[\Pkd(x^\top u)^2]=t^{-2}\Nkd^{-1},
	\end{align}
	which proves the desired result.
\end{proof}

\begin{lemma}[Lemma 1 of \citet{laurent2000adaptive}]\label{lem:laurent-massart} 
	Let $a_1, \cdots, a_d$ be i.i.d. $\calN(0,1)$ Gaussian variables. Then for any $t>0$,
	\begin{align}
		\Pr\(\sum_{i=1}^{d}a_i^2\le d-2\sqrt{dt}\)\le \exp(-t).
	\end{align}
\end{lemma}
\begin{lemma}\label{lem:gaussian-proj-concentration}
	Let $a=(a_1, \cdots, a_d)\sim \calN(0,I)$ be a Gaussian vector and $u\in\R^{d}$ a fixed vector with unit norm. Then for any $t>0$,
	\begin{align}
		\Pr\(\frac{|\dotp{a}{u}|}{\|a\|_2}\ge \frac{2t}{\sqrt{d}}\)\le 3\exp(-t^2/2).
	\end{align}
\end{lemma}
\begin{proof}
	Since $a\sim \calN(0,I)$ is a Gaussian vector, we have $\dotp{a}{u}\sim \calN(0,1).$ Hence,
	\begin{align}
		\Pr\(|\dotp{a}{u}|>t\)\le 2\exp(-t^2/2).
	\end{align}
	By Lemma~\ref{lem:laurent-massart} with $t=\frac{9d}{64}$, we also have
	\begin{align}
		\Pr\(\|a\|_2\le \sqrt{d}/2\)=\Pr\(\|a\|_2^2\le d/4\)\le \exp\(-\frac{9}{64}d\)\le \exp\(-\frac{1}{8}d\).
	\end{align}
	By union bound we have
	\begin{align}
		&\Pr\(\frac{|\dotp{a}{u}|}{\|a\|_2}\ge \frac{2t}{\sqrt{d}}\)\le \Pr\(|\dotp{a}{u}|>t\)+\Pr\(\|a\|_2\le \sqrt{d}/2\)\\
		\le\;&2\exp(-t^2/2)+\exp\(-\frac{1}{8}d\).
	\end{align}
	Note that when $t>\sqrt{d}/2$, the desired result is trivial because $|\dotp{a}{u}|\le \|a\|_2$ with probability 1. Therefore, when $t\le \sqrt{d}/2$ we have
	\begin{align}
		\Pr\(\frac{|\dotp{a}{u}|}{\|a\|_2}\ge \frac{2t}{\sqrt{d}}\)\le 2\exp(-t^2/2)+\exp\(-\frac{1}{8}d\)\le 
		3\exp(-t^2/2).
	\end{align}
\end{proof}

\begin{lemma}\label{lem:nn-sample}
	Let $f$ be a infinite-width two-layer neural network with activation $\sigma:[-1,1]\to [-1,1]$, defined by 
	\begin{align}
		f(x)=\E_{w\sim \Sp}[\sigma(x^\top w)c(w)]
	\end{align}
	for some weight $c:\Sp\to\R$ with $\|c\|_1<\infty.$ For any $\epsilon>0$, there exists a neural network $\hat{f}$ with $m=4\|c\|_1^2/\epsilon^2$ neurons, defined by $\hat{f}(x)=\sum_{j=1}^{m}a_i\sigma(w_i^\top x)$, such that 
	$\|\hat{f}-f\|_2\le \epsilon$, and $\sum_{j=1}^{m}|a_i|\le \|c\|_1.$
\end{lemma}
\begin{proof}
	We prove this theorem by probabilistic method. Let $p:\Sp\to\R_+$ be a function given by $p(w)=|c(w)|/\|c\|_1$. Then $p$ is a probability density function. Let $m=4\|c\|_1^2/\epsilon^2$. We sample $w_1,\cdots,w_m$ independently from $p$ and let $a_i=\sign(c(w_i))\frac{\|c\|_1}{m}.$ Define the two-layer neural network $\hat{f}$ by $\hat{f}(x)\defeq \sum_{j=1}^{m}a_i\sigma(w_i^\top x).$ In the following we prove that $\E[\|\hat{f}-f\|_2^2]\le \epsilon^2$ where the expectation is taken over the random variables $w_1\cdots,w_m$.
	
	For any fixed $x\in\Sp$, we have
	\begin{align}
		&\hat{f}(x)-f(x)=\sum_{j=1}^{m}a_i\sigma(w_i^\top x)-f(x)
		=\frac{\|c\|_1}{m}\sum_{j=1}^{m}\(\sign(c(w_i))\sigma(w_i^\top x)-\frac{f(x)}{\|c\|_1}\).
	\end{align}
	Let $X_i\defeq \sign(c(w_i))\sigma(w_i^\top x)-\frac{f(x)}{\|c\|_1}.$ By basic algebra we have
	\begin{align}
		\E_{w_i}[X_i]&=\int_{\Sp} p(w_i)\sign(c(w_i))\sigma(w_i^\top x)\dd w_i - \frac{f(x)}{\|c\|_1}\\
		&=\frac{1}{\|c\|_1}\(\int_{\Sp} |c(w_i)|\sign(c(w_i))\sigma(w_i^\top x)\dd w_i - f(x)\)\\
		&=\frac{1}{\|c\|_1}\(\int_{\Sp} c(w_i)\sigma(w_i^\top x)\dd w_i - f(x)\)=0.
	\end{align}
	In addition, $|X_i|\le |\sign(c(w_i))\sigma(w_i^\top x)|+|\frac{f(x)}{\|c\|_1}|\le 2.$ It follows that
	\begin{align}
		\E_{\hat{f}}[(\hat{f}(x)-f(x))^2]&=\frac{\|c\|_1^2}{m^2}\(\sum_{j=1}^{m}X_i\)^2=\frac{\|c\|_1^2}{m^2}\sum_{j=1}^{m}X_i^2\le \frac{4\|c\|_1^2}{m}.
	\end{align}
	Consequently,
	\begin{align}
		\E_{\hat{f}}[\|\hat{f}-f\|_2^2]=\E_{\hat{f}}[\E_{x\in\Sp}[\|\hat{f}-f\|_2^2]]\le \frac{4\|c\|_1^2}{m}\le \epsilon^2.
	\end{align}
	By the probabilistic method, there exists $\hat{f}$ such that $\|\hat{f}-f\|_2^2\le \epsilon^2$, which proves the first part of the lemma.
	
	By construction, we also have 
	\begin{align}
		\sum_{j=1}^{m}|a_j|=\sum_{j=1}^{m}\frac{\|c\|_1}{m}\le \|c\|_1
	\end{align}
	almost surely, which proves the second part of the lemma.
\end{proof}

\begin{proposition}\label{prop:truncation-upperbound}
	For any fixed $\alpha\in (0,1],c_1,c_2>0,\epsilon>0$, let $$k=\inf_{l\ge 0}\{2c_1c_2(l+1)^{3/2}(N_{l+1,d})^{-\alpha/2}\le \epsilon/2\}.$$ When $d> \max\{2e,4/\alpha\}$ we have $k\le \max\{2e,(4c_1c_2/\epsilon)^{\frac{2}{d\alpha-3}}\}$ and $\Nkd\le \(\frac{4c_1c_2}{\epsilon}\)^{8/\alpha}$.
\end{proposition}
\begin{proof}
	Let $c=c_1c_2$. By the definition $k$ we get
	\begin{align}\label{equ:tu-1}
		2c(k+1)^{3/2}N_{k+1,d}^{-\alpha/2}\le \epsilon/2.
	\end{align}
	Consequently, by the fact that $N_{k+1,d}={{d+k}\choose {d-1}} - {{d+k-2}\choose {d-1}}\le {d+k+1 \choose d}\le \(\frac{e(d+k+1)}{d}\)^{d}$ we get
	\begin{align}
		4c(k+1)^{3/2}\le \epsilon N_{k+1,d}^{\alpha/2}\le \epsilon \(\frac{e(k+d+1)}{d}\)^{d\alpha/2}.
	\end{align}
	When $d\ge 2e$ and $k\ge 2e$, we get $\frac{e(k+d+1)}{d}\le k+1.$
	As a result,
	\begin{align}
		4c(k+1)^{3/2}\le \epsilon (k+1)^{d\alpha/2},
	\end{align}
	which implies that 
	\begin{align}
		k\le \(\frac{\epsilon}{4c_1c_2}\)^{\frac{2}{3-d\alpha}}=\(\frac{4c_1c_2}{\epsilon}\)^{\frac{2}{d\alpha-3}}.
	\end{align}
	By the definition $k$ we also have
	\begin{align}
		2ck^{3/2}N_{k,d}^{-\alpha/2}> \epsilon/2.
	\end{align}
	Hence, 
	\begin{align}
		\Nkd\le \(\frac{4c}{\epsilon}\)^{2/\alpha}k^{3/\alpha}\le \(\frac{4c}{\epsilon}\)^{8/\alpha}
	\end{align}
\end{proof}

\begin{proposition}\label{prop:lipschitzness-poly}
	For any $k\ge 0$, let $\sigma_k(t)=\sum_{l=0}^{k}\dotp{\relu}{\bar{P}_{l,d}}_\mud \bar{P}_{l,d}(t)$. Then for all $k\ge 0$ we have
	\begin{align}
		&\sup_{t\in[-1,1]}|\sigma_k(t)|\le 1200\sqrt{\Nkd},\\
		&\sup_{t\in[-1,1]}|\sigma_k'(t)|\le 1200k\sqrt{\Nkd}.
	\end{align}
\end{proposition}
\begin{proof}
	Let $\tau_l=\dotp{\relu}{\bar{P}_{l,d}}_\mud.$ Then we have
	\begin{align}
		\sup_{t\in[-1,1]}|\sigma_k(t)|\le \sum_{l=0}^{k}\tau_l \sup_{t\in[-1,1]}|\Pldb(t)|=\sum_{l=0}^{k}\tau_l\sqrt{\Nld}\le k\tau_k\sqrt{\Nkd}.
	\end{align}
	By Lemma~\ref{lem:relu-eigen-approx} we get $l\tau_l\le 1200.$ As a result,
	\begin{align}
		\sup_{t\in[-1,1]}|\sigma_l(t)|\le 1200\sqrt{\Nld}.
	\end{align}

	By \citet[Eq. (2.89)]{atkinson2012spherical} we have $\sup_{t\in[-1,1]}|\Pldb'(t)|\le \frac{l(l+d-2)}{d-1}.$ As a result,
	\begin{align}
		&\sup_{t\in[-1,1]}|\sigma_l'(t)|\le \sum_{l=0}^{k}\tau_l \sup_{t\in[-1,1]}|\Pldb'(t)|=\sum_{l=0}^{k}\tau_l\frac{l(l+d-2)}{d-1}\sqrt{\Nld}\\
		\le\;& \tau_k\frac{k^2(k+d-2)}{d-1}\sqrt{\Nkd}\le 1200k\sqrt{\Nkd}.
	\end{align}
\end{proof}

\end{document}